\documentclass[lettersize,journal]{IEEEtran}
\usepackage{graphics} 
\usepackage{epsfig} 
\usepackage{times} 
\usepackage{amsfonts,amssymb,amsmath,latexsym}
\usepackage{multirow,multicol}
\usepackage[ruled,linesnumbered]{algorithm2e}
\usepackage{algpseudocode}
\usepackage{hyperref}
\usepackage{epstopdf} 
\usepackage{subfigure} 
\usepackage{stfloats} 
\usepackage{bm}
\usepackage{color}
\usepackage{amsthm}
\usepackage[square,comma,numbers,sort&compress]{natbib}
\usepackage{array}
\usepackage{soul}
\usepackage{threeparttable}  
\usepackage{hyperref}

\hypersetup{
    colorlinks=true,
    urlcolor=blue,
    citecolor=blue,
    linkcolor=blue,
}
\begin{document}
\title{Discretizing Continuous Action Space with Unimodal Probability Distributions for On-Policy Reinforcement Learning}
\author{Yuanyang~Zhu, Zhi~Wang,~\IEEEmembership{Member,~IEEE}, Yuanheng~Zhu,~\IEEEmembership{Senior Member,~IEEE}, Chunlin~Chen,~\IEEEmembership{Senior Member,~IEEE}, Dongbin~Zhao,~\IEEEmembership{Fellow,~IEEE}

\thanks{The work was supported by the National Natural Science Foundation
of China under Grant 62006111 and Grant 62073160.}
	
\thanks{Yuanyang Zhu, Zhi Wang and Chunlin Chen are with the Department of Control Science and Intelligent Engineering, School of Management and Engineering, Nanjing University, Nanjing 210093, China, and also with the Research Center for Novel Technology of Intelligent Equipment, Nanjing University, Nanjing 210093, China (e-mail: yuanyang@smail.nju.edu.cn; zhiwang@nju.edu.cn; clchen@nju.edu.cn).}
 
 \thanks{Yuanheng Zhu and Dongbin Zhao are with the State Key Laboratory of Management and Control for Complex Systems, Institute of Automation, Chinese Academy of Sciences, Beijing 100190, China, and also with the School of Artificial Intelligence, University of Chinese Academy of Sciences, Beijing 100049, China (e-mail: yuanheng.zhu@ia.ac.cn; dongbin.zhao@ia.ac.cn)}
}

\markboth{IEEE Transactions on Neural Networks and Learning Systems}%
{Shell \MakeLowercase{\textit{et al.}}: Bare Demo of IEEEtran.cls for Computer Society Journals}

\IEEEtitleabstractindextext{
\begin{abstract}
For on-policy reinforcement learning, discretizing action space for continuous control can easily express multiple modes and is straightforward to optimize.
However, without considering the inherent ordering between the discrete atomic actions, the explosion in the number of discrete actions can possess undesired properties and induce a higher variance for the policy gradient estimator. 
In this paper, we introduce a straightforward architecture that addresses this issue by constraining the discrete policy to be unimodal using Poisson probability distributions.
This unimodal architecture can better leverage the continuity in the underlying continuous action space using explicit unimodal probability distributions.
We conduct extensive experiments to show that the discrete policy with the unimodal probability distribution provides significantly faster convergence and higher performance for on-policy reinforcement learning algorithms in challenging control tasks, especially in highly complex tasks such as Humanoid.
We provide theoretical analysis on the variance of the policy gradient estimator, which suggests that our attentively designed unimodal discrete policy can retain a lower variance and yield a stable learning process.
\end{abstract}

\begin{IEEEkeywords}
Reinforcement learning, unimodal probability distributions, Poisson distributions, on-policy learning
\end{IEEEkeywords}}

\maketitle

\IEEEdisplaynontitleabstractindextext

\IEEEpeerreviewmaketitle

\section{Introduction}\label{sec:introduction}
\begin{figure}[tb]
\begin{center}
\setlength{\belowcaptionskip}{-5mm}
\centerline{\includegraphics[width=0.95\columnwidth]{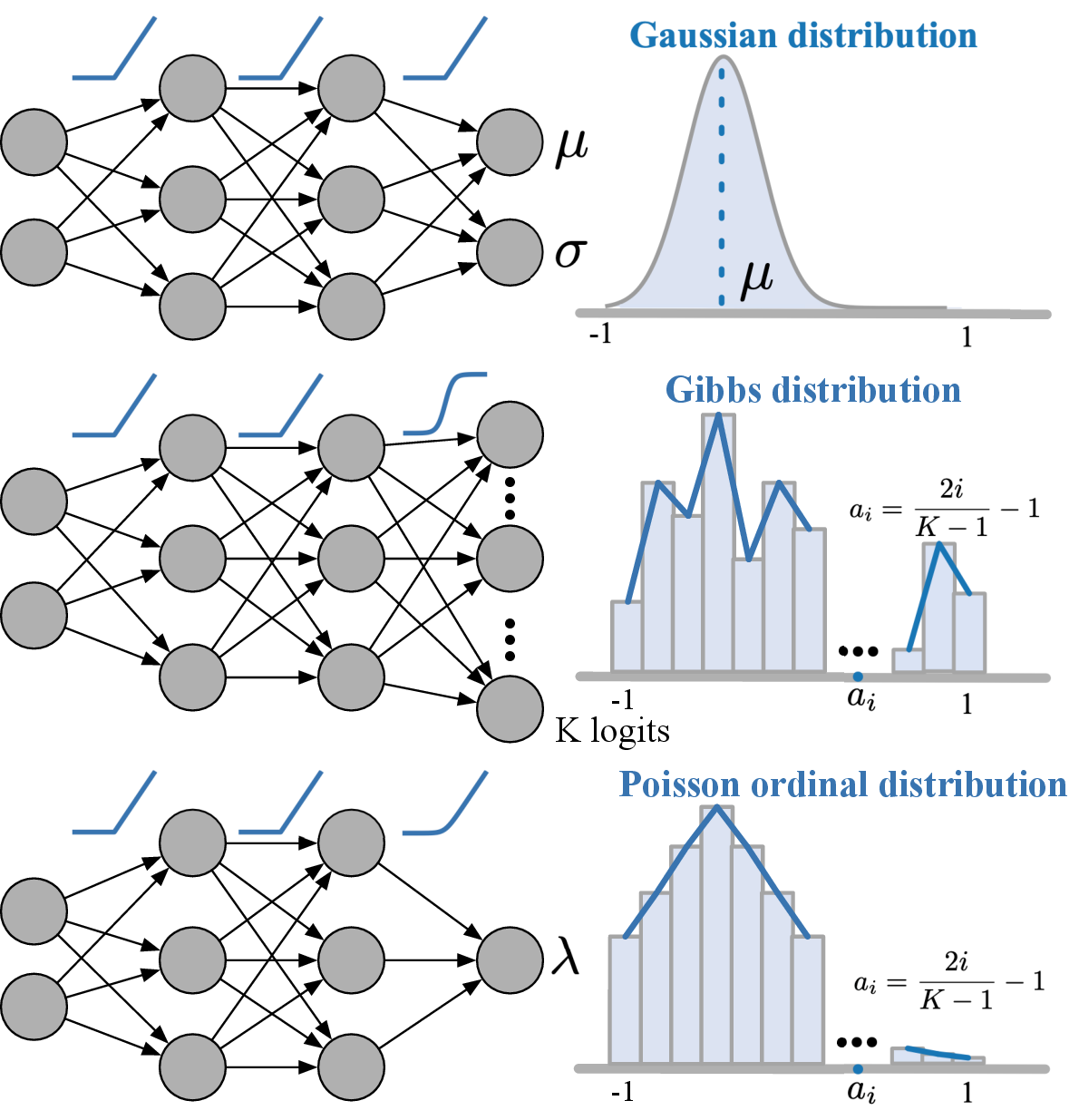}}
\caption{An example of continuous control with different action probability distributions over only one action dimension.
Top: The unimodal continuous policy distribution is constructed from the Gaussian distribution that estimates its mean $\mu$ and standard deviation $\sigma$ with a function approximator such as deep neural networks.
Middle: When there are $K$ discrete actions, the policy distribution can be represented by a Gibbs distribution, wherein $K$ logits are generated by a function approximator via a sigmoid function.
This results in a distribution that displays multi-modality characteristics.
Bottom: When there are $K$ discrete actions, the new unimodal ordinal policy distribution can be characterized by the Poisson ordinal distribution, which outputs a probability mass function ($\lambda$) via a function approximator through a Softplus function. 
This distribution ensures that the two classes adjacent to the majority class receive the next greatest probability mass.
}
\vskip -0.28in
\label{distribution}
\end{center}
\end{figure}
\IEEEPARstart{D}{eep} reinforcement learning (DRL) has presented a powerful paradigm for learning new behaviors from scratch both on physical and simulated challenging tasks~\cite{zhou2023learning,haarnojalearning,peng2020learning,10021988}.
The action space of conventional reinforcement learning (RL) tasks can be discrete, continuous, or some combination of both~\cite{gu2016continuous,haarnoja2017reinforcement}.
In simulation~\cite{schulman2015trust} and real life~\cite{levine2016end, riedmiller2018learning, andrychowicz2020learning,yang2018hierarchical}, continuous control usually requires some subtle parametric functions for a compact representation of action distributions, typically Gaussians, due to an infinite number of feasible actions in a continuous action space~\cite{3272068,lillicrap2015continuous,yang2018hierarchical,9511019}.
This underlying assumption is that it can enable more refined decisions when control policies can cover all feasible control inputs, which brings difficulty in maximizing an arbitrary function for a continuous action space~\cite{9353402,dadashi2021continuous}.
Furthermore, it hinders applying dynamic programming approaches to back up value function estimates from successor states to parenting states.

Discretizing a continuous control problem is a scalable solution that retains the simplicity of discrete actions and allows for theoretical analysis~\cite{andrychowicz2020learning,seyde2021bang,sun2023}.
With such a setup, impressive results have been obtained in high-dimensional continuous action spaces~\cite{andrychowicz2020learning,jaskowski2018reinforcement,tang2020discretizing}.
However, the naive discretization approach results in exponentially large discrete action space, where ballooning action space can quickly become intractable.
It would be a balancing dilemma that the resulting output may not be smooth enough when the discretization is coarse, and the number of discretized actions may be highly intractable otherwise.
Moreover, more fine discretization of the action space may struggle to capture the information about the class ordering of the continuous action space, making it challenging to generalize across discrete actions.
Furthermore, it also brings a larger variance of the policy gradient estimator, which may hurt the stability and performance, especially for highly complex tasks~\cite{song2019v,wu2018variance}. 
In practice, more fine-tuned hyperparameters are also required to prevent policy collapse.

To tackle these intractabilities, prior works~\cite{tang2020discretizing, wu2021learning} assume that action dimensions are independent and developed to factorize distributions across action dimensions using an ordinal parameterization of the joint distribution over discrete actions, which has shown improved performance in finding near-optimal solutions for high-dimensional tasks.
By parameterizing the discrete policy using an ordinal parameterization, the natural ordering between discrete actions can be implicitly encoded, which provides an additional inductive bias that can improve the generalization across actions~\cite{chu2005gaussian, niu2016ordinal, liu2019unimodal}.
In contrast, the conventional approach of parameterizing the discrete policy as a Gibbs distribution fails to consider the order of actions and ignores the discrepancy relationships between them~\cite{beckham2017unimodal}.
Existing methods usually use some heuristic network architecture design to implicitly encode the ordinal information into the distribution over discrete actions, which may not guarantee a unimodal probability distribution to efficiently capture the ordinal information within atomic actions~\cite{cheng2008neural}.
In their work, they alleviate this issue by designing the loss function to impose an ordering on the action, which does not impose unimodal constraints.
Here, we propose a more explicit and efficient ordinal parameterization method that better incorporates the notion of continuity when parameterizing the distribution over discrete actions, while avoiding the need for an exponentially large number of actions. 
Additionally, we impose an ordering constraint on the action space of each dimension through the loss function to further improve the performance of RL algorithms.

This paper mainly addresses how to efficiently leverage the internal ordering information in the continuous action space to enhance generalization across the discretized set of actions for on-policy RL without an explosion in the number of parameters.
Inspired by the deep ordinal classification approach~\cite{beckham2017unimodal}, for each independent action dimension, we constrain the distribution over discrete actions to be unimodal to explicitly exploit the inherent ordering between discrete actions.
To implement the unimodal constraint, we employ the Poisson probability distribution, which allows the probability mass to decrease gradually on both sides of the action with the majority of the mass.
Our framework necessitates learning a probability mass function for each action dimension, substantially reducing the number of units in the network output layer and enabling us to circumvent the curse of dimensionality as the discretization action bins increase (see Fig.~\ref{distribution}).
It facilitates fine-grained discretization of individual domains without incurring a significant increase in the number of parameters.
The experimental results show that our method outperforms the baselines by a wide margin in the suite of MuJoCo continuous control tasks~\cite{tunyasuvunakool2020dm_control}.

We highlight the following contributions of this work: 
\begin{itemize} 
  \item We propose a novel ordinal framework that enforces the distribution over discrete actions to be unimodal, effectively capitalizing on the continuity presented in the action space.
      We utilize the probability mass function of the Poisson distribution to enforce unimodal probability distributions for each dimension action space, which can place more confidence on the majority probability distribution beneﬁting from its inherent unimodal distribution.
  \item Theoretically, we present a variance analysis for the policy gradient estimator, where our unimodal policy attains a lower variance for the estimator.
  \item We show experimentally that an ordinal parameterization of the unimodal policy outperforms other competing approaches for on-policy RL on continuous control tasks.
  The promising results showcase the versatility and effectiveness of our methods compared to the baselines.
\end{itemize}

The remaining paper is organized as follows.
In Section~\ref{Sec2}, we summarize preliminaries of RL and related literature.
In Section~\ref{Sec3}, we present the unimodal probability distributions for on-policy RL.
In Section~\ref{Sec4}, we give comprehensive experimental results of learning performance and stability analysis.
The paper is concluded by Section~\ref{Sec5}.

\section{Background and related work}\label{Sec2}
\subsection{Preliminaries}
RL is an approach to solving optimal sequential decision-making tasks, built upon the concept of Markov Decision Processes (MDPs). 
In the finite MDP framework, a tuple of $\langle S,A,T,R,\gamma\rangle$ defines the problem, where $S$ is a countable state space, $A$ is a finite action set, $T: S \times A \times S \rightarrow[0,1]$ represents the transition kernel, ${R: S \times A} \to \mathbb{R}$ is the reward function, and $\gamma \in[0,1)$ is the discount factor. 
A stochastic policy is defined as $\pi: S \times A \rightarrow[0,1]$, which maps environmental states to distributions over actions with the constraint $\sum_{a\in {A}}\pi(a|s)=1,\forall s\in {S}$.
The objective of the RL problem is to find an optimal policy $\pi^{*}$ that maximizes the expected total discounted return by
\begin{equation}
J(\pi)=\mathbb{E}_{\tau \sim \pi(\tau)}[r(\tau)]=\mathbb{E}_{\tau \sim \pi(\tau)}\left[\sum_{i=0}^{\infty} \gamma^{i} r_{i}\right],
\end{equation}
\begin{equation}
Q_\pi(s, a)=\mathbb{E}_{s_0=s, a_0=a, \tau \sim \pi}\left[\sum_{t=0}^T \gamma^t r_t\right],
\end{equation}
where $\tau = (s_0, a_0, s_1, a_1, ...)$ is a learning episode, $\pi(\tau)=p(s_0)\Pi_{t=0}^{\infty}\pi(a_t|s_t)p(s_{t+1}|s_t,a_t)$, and $r_t$ is the immediate reward. 
The policy function $h(\cdot \mid \boldsymbol{\theta})$ can be represented by a deep neural network parameterized by $\theta$. 
For a discrete action space, the Gibbs distribution is commonly used and given by
\begin{equation}
\pi_{\theta}(a_j \mid s)=\frac{\exp \left(h_{j}(s \mid \boldsymbol{\theta})\right)}{\sum_{j \in A(s)} \exp \left(h_{j}(s \mid \boldsymbol{\theta})\right)},
\end{equation}
while the Gaussian distribution is generally used for a continuous action space and is given by
\begin{equation}
\pi_{\theta}(a \mid s)=\frac{1}{\sqrt{2 \pi} \sigma} \exp \left(-\frac{1}{\sigma^{2}}(h(s \mid \boldsymbol{\theta})-a)^{2}\right).
\end{equation}

To evaluate the performance of a policy $\pi_{\theta}$, the direct objective function is optimized by estimating the gradient of the expected return
\begin{equation}
J(\boldsymbol{\theta})=\mathbb{E}_{\tau \sim \pi_{\theta}(\tau)}[r(\tau)]=\int_{\tau} \pi_{\theta}(\tau) r(\tau) d \tau,
\end{equation}
where $r(\tau)=\sum_{i=0}^{\infty} \gamma^{i} r_{i}$ is the return of episode $\tau$.
The basic idea of the policy gradient algorithm~\cite{duan2016,sutton2018reinforcement} gives the direction of the performance gradient where the parameters $\theta$ should be updated:
\begin{equation}
\begin{aligned} \nabla_{\boldsymbol{\theta}} J(\boldsymbol{\theta}) &=\mathbb{E}_{\tau \sim \pi_{\theta}(\tau)}\left[\nabla_{\boldsymbol{\theta}} \log \pi_{\theta}(\tau) r(\tau)\right] \\ &=\int_{\tau} \nabla_{\boldsymbol{\theta}} \log \pi_{\theta}(\tau) r(\tau) \pi_{\theta}(\tau) d \tau \\ & \approx \sum_{i=1}^{m} \nabla_{\boldsymbol{\theta}} \log \pi_{\theta}\left(\tau^{i}\right) r\left(\tau^{i}\right), \end{aligned}
\end{equation}
where $\left(\tau^{1}, \ldots, \tau^{m}\right)$ is a batch of learning episodes sampled from policy $\pi_{\theta}$.
Next, an ascent step is taken in estimating the direction of the gradient as $\boldsymbol{\theta} \leftarrow \boldsymbol{\theta}+\alpha \nabla_{\boldsymbol{\theta}} J(\boldsymbol{\theta})$.
The process continues until $\theta$ converges~\cite{williams1992simple}.

\subsection{Related Work}
\textbf{On-policy RL.} On-policy RL is a family of RL algorithms aiming to directly optimize the parameters of a policy, which optimizes expected returns by estimating policy gradients.
This estimation is often prone to high variance, and several solutions have been proposed to mitigate this significant challenge, particularly in problems characterized by long horizons or high-dimensional action spaces~\cite{greensmith2004variance,schulman2015high, wu2018variance}.
Classic vanilla policy gradient (PG) updates are generally unstable with high variance due to difficulties in crediting actions that influence future rewards, where the gradients require knowledge of the probability of the performed action via resorting to parametric distributions.
Natural policy~\cite{kakade2001natural} improves upon vanilla PG by computing an ascent direction with the Fisher information that approximately ensures a slight change in the policy distribution.
To obtain more stable learning, trust region policy optimization (TRPO)~\cite{schulman2015trust} utilizes a larger Kullback-Leibler (KL) divergence value to enforce constraints, and performs a line search in the natural gradient direction, ensuring improvements in the surrogate loss function.
Proximal policy optimization (PPO)~\cite{schulman2017proximal} replaces the KL divergence constraint with the clipped surrogate objective, which strikes a favorable balance across sample complexity, simplicity, and wall time.
Moreover, actor-critic using Kronecker-factored trust region (ACKTR)~\cite{wu2017scalable} builds upon the natural policy gradient framework by calculating gradients through Kronecker-factored approximate curvature within the trust region.
Orthogonal to the above algorithms, we demonstrate that ordinal parameterization with a unimodal constraint for on-policy RL achieves consistently improved performance based on the above representative algorithms.

\textbf{Policy Representation.}
In the context of policy gradient techniques, the policy is generally parameterized using neural networks, where the policy is improved by optimizing the parameter of the approximation function.
In general, the policy uses a Gibbs probability distribution over the discrete action space to sample.
For the continuous control problems, the default choice for the policy in the baseline is parameterized by learning a Gaussian distribution with independent components for each dimensional action space~\cite{schulman2015trust, schulman2017proximal}.
The Gaussian mixture, maximum entropy, or normalizing ﬂows~\cite{rezende2015variational,haarnoja2018soft,tang2018implicit} can be used for more expressive policy classes, providing a promising avenue for improved robustness and stability.
Since physical constraints in most continuous control tasks, actions can only take on values within some ﬁnite interval, resulting in an unavoidable estimation bias caused by boundary effects.
To address the shortcomings of the Gaussian distribution with a finite support distribution, the Beta distribution~\cite{chou2017improving} provides a bias-free and less noise policy with improved performance.
Further, considering the extremes along each action dimension, a Bernoulli distribution is applied to replace the Gaussian parametrization of continuous control methods and show improved performance on several continuous control benchmarks~\cite{seyde2021bang}.
Here, we provide insight into how to produce an efficient ordinal parameterization policy while retaining the internal order information of continuous action space with a simple explicit constraint, which can yield state-of-the-art performance compared to baseline policy classes.

\textbf{Continuous action discretization.}
To leverage the continuity between discrete and continuous action space, converting continuous control problems into discrete ones has been introduced by~\cite{bushaw1953differential} with the ``bang-bang" controller~\cite{bellman1956bang}.
Similar discretization methods represent each action in binary format and optimize its policy for MDPs with large action sets~\cite{sallans2004reinforcement,pazis2011generalized,cui2016online,vieillard22a,tavakolilearning}.
However, such discretization methods need to be improved due to the curse of dimensionality.
Surprisingly, recent works have relieved the limitation by assuming that action dimensions are independent~\cite{7d4c0094,tang2020discretizing,andrychowicz2020learning,metz2017discrete,tavakoli2020learning}, which obtains improved performance in complex control tasks.
Inspired by multi-agent reinforcement learning, Decoupled Q-Networks (DecQN)~\cite{seyde2022solving} factorizes the overall state-action value function into a linear combination of single action utility functions over action dimensions by combining value decomposition with bang-bang action space discretization.
It achieves better performance while reducing agent action space complexity.
However, such a discretization paradigm makes the strong assumption of independence among each action dimension and hardly captures the continuous order information of actions~\cite{tavakoli2018action,sakryukin2020inferring}.

To mitigate the exponential explosion of discrete actions,~\cite{metz2017discrete} utilizes sequence-to-sequence models to develop policies for structured prediction problems, but their strategy has only demonstrated effectiveness in high-dimensional tasks like Humanoid and HumanoidStandup tasks.
Recently,~\cite{tang2020discretizing} parameterizes the discrete distributions with an ordinal architecture and achieves improved performance.
Nevertheless, existing methods face a dilemma in that the number of network outputs increases linearly with the number of action space dimensions, which may bring high variance, resulting in a more unstable learning process.
A key distinction between our work and previous works lies in our explicit utilization of unimodal distributions with only one parameter for each action dimension. 
It can help the policy capture the most confident class while ensuring that the probability gradually decreases on both sides of the class. 
Finally, we find that our proposed method outperforms baseline algorithms, especially in high-dimensional tasks, which should benefit from the unimodal distribution architecture.
We summarize the above continuous action discretization RL works in Table~\ref{summary}.

\begin{table}[tb]
\caption{Summary of continuous action discretization and ours regarding the continuity of the discretized action bins, the modal of policy distribution, the combination complexity of the action space, and the independence of each dimension action space.
$K$ and $N$ denote the discretized action bins for each action dimension and the action dimension, respectively.}

\begin{center}
\scriptsize
\begin{tabular}{lcllcl}
\hline \hline Method  & Continuity & Modal & Complexity & Independence \\
\hline \cite{bushaw1953differential,bellman1956bang}  & No & Multimodal & $K^N$ & Yes  \\
\cite{sallans2004reinforcement,cui2016online}  & No & Multimodal & $K^N$ & No \\
\cite{pazis2011generalized}  & No & Multimodal & $K^N$ & Yes \\
\cite{vieillard22a,tavakolilearning,tang2020discretizing,andrychowicz2020learning} & Yes & Multimodal & $KN$ & Yes \\
\cite{metz2017discrete,sakryukin2020inferring}  & No & Multimodal & $KN$ & No \\
\cite{tavakoli2018action}  & No & Multimodal & $KN$ & Yes \\
Ours  & Yes & Unimodal & $KN$ & Yes \\
\hline \hline
\end{tabular}
\end{center}
\label{summary}
\end{table}

\section{Unimodal Probability Distributions for On-Policy Reinforcement Learning}\label{Sec3}
In this section, we will present our proposed unimodal distribution with Poisson probability distribution for on-policy RL algorithms, which explicitly introduce unimodal probability distribution into the ordinal parameterization method to leverage the continuity in the underlying continuous action space.
First, we present the process of discretizing action space for continuous control tasks and apply it to on-policy RL.
Then, we describe the unimodal ordinal architecture that constrains the discrete policy to be unimodal via the Poisson probability distribution in a practical neural network implementation.
Finally, theoretical analysis shows that our method can relieve the high variance trouble of the policy gradient estimator compared to existing ordinal parameterization methods.

\subsection{Discretizing Action Space for Continuous Control}
Consider an environment with state $s_t$ and $m$ dimension action spaces $a \in \mathbb{R}^m$.
To ensure generality, we consider an action space $\mathcal{A}=[-1,1]^m$ and proceed to discretize each dimension of the action space into $K$ evenly spaced atomic action bins.
As a result, we obtain a discrete and finite set of atomic actions $\mathcal{A}_{i}=\{\frac{2 j}{K-1}-1 \}_{j=0}^{K-1}$ for the $i$ action dimension. 
To facilitate the joint policy's tractability, the policy generates a tuple of discrete actions accompanied by factorized categorical distributions across action dimensions. 
This approach obviates the need to enumerate every possible actions, as required in the discrete case.
Specifically, we denote a categorical distribution $\pi_{\theta_{i} ({a_j} \mid {s})}$ as the atomic action for actions $a_{i} \in {\mathcal{A}_{i}}$ at each step $t$, where $\theta_{i}$ is the factorization parameters for this marginal distribution induced by Poission probability distributions.
Such representations can build complex policy distributions with explicit probability distributions, where any action subset can be formulated as
\begin{equation}
	\pi \left(\bm{{a}} \mid {s}\right) = \Pi_{i=1}^{m} \pi_{\theta _i} \left(a_{i} \mid {s}\right),
\end{equation}
where $\bm{a} = [{a_0, a_1, ..., a_{K-1}}]^{T} $ and $\pi_{\theta _i} \left(a_{i} \mid {s}\right)$ represent the probabilities of selecting the actions $a_i$ for the $i$-th dimension action space.
Based on this factorization, we can easily employ RL algorithms to maintain a tractable distribution over joint actions. 
Meanwhile, it can vastly reduce the computation of neural networks due to the fewer learning parameters.

\subsection{Unimodal ordinal architecture}
In on-policy RL algorithms, the policy gradient involves a stochastic policy that specifies the probability distribution of selecting an action $\bm{a}$ given a state $\bm{s}$.
Upon discretizing the continuous action space, the complexity of directly training such policy networks could lead to $K^m$ combinations of joint atomic actions, which exponentially increase with the growing number of action dimensions $m$.
Using the categorical distribution (e.g., Gumbel-Softmax distribution) as a stochastic policy for discrete action space has been well-studied in the RL community.
Popular categorical distributions have been used as a stochastic policy for discrete action spaces due to their ease of sampling and computation.
However, the curse of dimensionality can lead to intractable action spaces, putting significant pressure on the distribution parameterization over discrete actions.
In fact, a fine control policy can be captured more as the resolution of the discretized action space increases. 
This quickly plunges into the dilemma that intractable action space puts high pressure on the parameterizing distributions over discrete actions, which hinders capturing the ordering among actions.

To relieve the scalability issue, a few prior works have been exploring more expressive distributions and notably proposed an ordinal parameterization~\cite{tavakoli2018action,tang2020discretizing,sakryukin2020inferring}.
It introduces an implicit loss function based on factorizing a tractable distribution over joint actions, where the ordinal policy can easily do both sampling and training.
Motivated from the loss function of~\cite{cheng2008neural}, it implicitly introduces internal ordering information between classes while maintaining the probabilistic properties of discrete distributions instead of simply parameterizing the discrete policy as a categorical distribution, ignoring the order of actions and the discrepancy relationship between the joint actions.
Existing methods usually utilize some heuristic design of the network architecture to learn distributions for each action dimension, which could not guarantee a unimodal probability distribution output followed by the neural network to efficiently capture the ordinal information within atomic actions.
It generally has a strong underlying assumption that the output logits of the neural network are unimodal for each dimension action space, which makes it hard to guide the neural network to optimize weights and generate probability distributions that closely resemble the optimal action.
In other words, it may allocate greater confidence to multi-action classes, as it generally produces multi-modal probability distribution.

To leverage the continuity in the underlying action space, we focus on how the discretized actions are distributed: why restrict ourselves to the categorical distribution?
For continuous control tasks, a Gaussian or Beta policy can explicitly capture the inherent ordering structure within an infinite number of actions due to the privilege of their unimodal distribution property.
Is it possible to explicitly retain the unimodal structure and inherent ordinal information after discretizing the continuous action space for learning a policy over discrete actions?
Inspired by the deep ordinal classification approach~\cite{da2008unimodal,beckham2017unimodal}, we constrain the distribution of the discretized actions to be unimodal explicitly for utilizing the ordinal information about the underlying continuous space.
Specifically, we learn the probability mass function (PMF) of the Poisson distribution to enforce discrete unimodal probability distributions for each action dimension.
This PMF can facilitate the generation of the most coherent unimodal probability distribution, as it diminishes the probability mass on both sides of the action possessing the majority of the mass, allowing us to adeptly capture the intrinsic ordering structure present within the action space.

The Poisson distribution, a widely recognized discrete probability distribution, is typically used to model the likelihood of observing a specific count of events within a designated time interval.
It is characterized by a positive constant mean rate, denoted as $\lambda \in R^{+}$, and presumes that events transpire independently of the elapsed time since the previous event. 
In other words, the distribution describes the probability of $k \in \mathbb{N}$ events occurring during a given time interval, where $k$ can take on any non-negative integer value.
For $i$-th dimension of action space, the Poisson distribution function is denoted as $p(k; \lambda_i)=\frac{\lambda^{k} \exp (-\lambda_i)}{k !}$, and $0 \leq k \leq {K-1}$ (minus one since we index from zero).
For a purely technical implementation reason, we focus on the log of the PMF, which can be expressed as
\begin{equation}
\label{equ_log}
\begin{aligned}
\log \left[\frac{\lambda_i^{k} \exp (-\lambda_i)}{k !}\right] &=\log \left(\lambda_i^{k} \exp (-\lambda_i)\right)-\log (k !) \\
&=\log \left(\lambda_i^{k}\right)+\log (\exp (-\lambda_i))-\log (k !) \\
&=k \log (\lambda_i)-\lambda_i-\log (k !).
\end{aligned}
\end{equation}
We denote the scalar output of our deep network as $F(\bm{s}) = [f_1(\bm{s}), f_2(\bm{s}),..., f_i(\bm{s})]$, where $f_i(\bm{s})>0$ is enforced to be positive using the Softplus nonlinearity.
With such a parameterization, our method only needs to learn $M$ probability mass functions with $M$ units in the network output layer, in contrast to existing ordinal parameterization methods that require learning the complete probability distributions over all dimensions using $M*K$ units in the network output layer~\cite{tang2020discretizing}.
Especially in high-dimensional continuous action spaces, our unimodal method can make it more tractable to implement with deep neural network representations.
Meanwhile, our method can eliminate the assumption that the ordinal parameterization method could produce a unimodal probability distribution only when the logits of the network output are unimodal.
By simply replacing the $\lambda_i$ in equation~(\ref{equ_log}) with $f_i(\bm{s})$, we denote $h_i(\bm{s})_j$ to be
\begin{equation}
j \log (f_i(\mathbf{s}))-f_i(\mathbf{s})-\log (j !).
\end{equation}

\begin{figure}[tb]
\begin{center}
\setlength{\belowcaptionskip}{-5mm}
\centerline{\includegraphics[width=0.99\columnwidth]{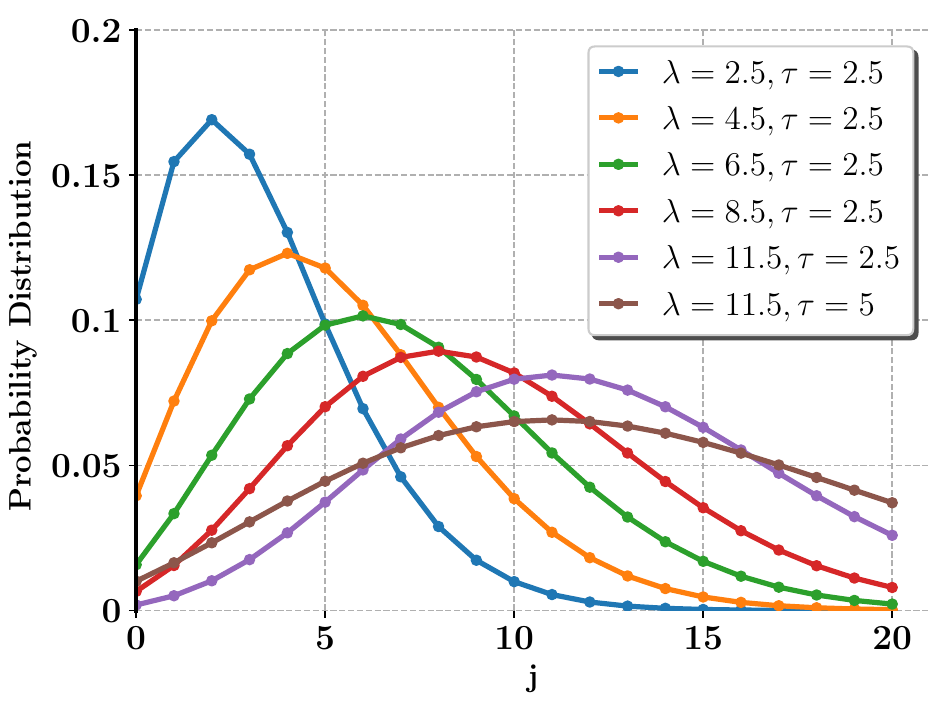}}
\caption{Normalization of the log-likelihood Poisson distributions. 
For each curve, we sample $21$ action distributions $j\in [0,20]$ and plot the normalized log-likelihood of the Poisson distribution curve with different values of the network output $f(s)$ and the temperature $\tau$ by evaluating $p(a_{ij}\mid s)$ with the Eq.~\ref{softmax}.
The maximum probability will peak at the $f(s)$, where the probability mass gradually decreases on both sides of the class.}
\vskip -0.25in
\label{Poisson}
\end{center}
\end{figure}

\begin{figure}[tb]
\begin{center}
\centerline{\includegraphics[width=0.99\columnwidth]{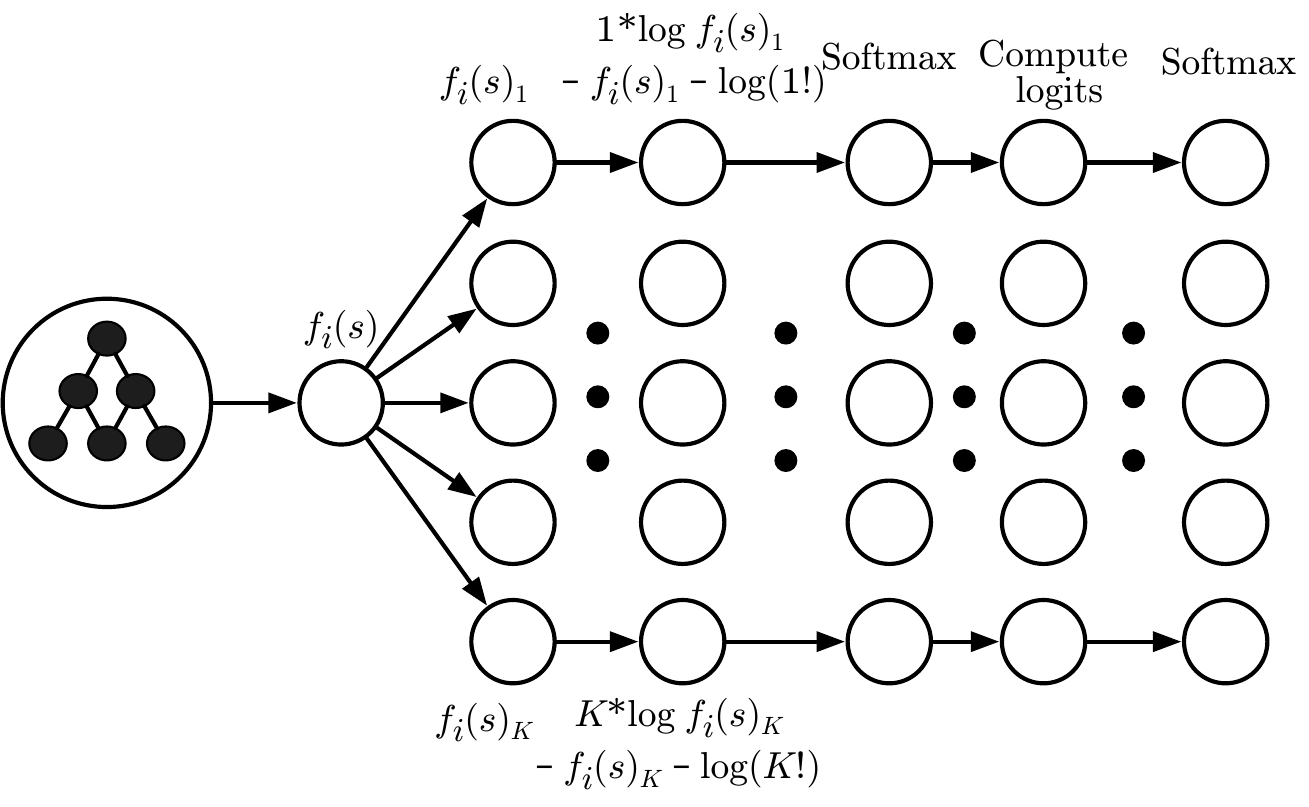}}
\caption{For simplicity, we illustrate the operation of the unimodal distribution over only one action dimension $i$.
the first layer following $f_i(x)$ acts as a `copy' layer, where $f_i(s)=f_i(s)_1=...=f_i(s)_K$.
The second layer applies the log Poisson PMF transform, followed by the Softmax layer.
The third layer normalizes the required probability distributions since the support of the Poisson is infinite.
We then compute the final logits by ordinal parameterization as in Eq.~\ref{logits}.
Finally, we derive the final output probability via a Softmax operation, where the actions are sampled according to this output distribution.}
\vskip -0.2in
\label{frame}
\end{center}
\end{figure}

In the RL community, the categorical distribution architecture is widely used to parameter a discrete distribution over $K$ action classes, which is represented by $K$ logits denoted as $h_i(\mathbf{s})_j$.
Due to the fact that the Poisson distribution has infinite support, we use a `right-truncated' operation by applying a Softmax operation and obtain the corresponding probability distribution $p(a_{ij})$ as
\begin{equation}
p(a_{ij} \mid \bm{s})=\frac{\exp \left(-h_i(\bm{s})_{j}/ \tau \right)}{\sum_{j=0}^{K-1} \exp \left(-h_i(\bm{s})_{j}/ \tau \right)}, 0\leq j \leq K-1.
\label{softmax}
\end{equation}
Besides, we introduce the hyperparameter $\tau$ to the Softmax function to control the variance of the distribution.\footnote{Note that as $\tau \rightarrow \infty$, the probability distribution becomes more uniform, leading to more efficient exploration. Conversely, the distribution becomes more `one-hot' like encouraging more deterministic behavior otherwise.}
After these operations, the probability distribution remains unimodal, which is clearly illustrated in Fig.~\ref{Poisson}.

In our unimodal policy architecture, we retain these logits $h_i(\bm{s})_j$ with the support of the Poisson distribution, and then transform them via a Softmax operation to relieve the infinite support problem of Poisson distribution.
Inspired by the ordinal architecture introduced in~\cite{tavakoli2018action,tang2020discretizing}, in order to enhance the additional dependencies between logits $h_i(\bm{s})_j$, we use the ordinal architecture to implicitly inject the information about the action ordering while maintaining all the probabilistic properties of Poisson distribution.
The ability to readily distinguish samples drawn from action $k$ from those drawn from all actions $j>k$ is advantageous, as it facilitates the identification of maxima in such cases. 
Therefore, for all logits $\forall 1 \leq l \leq K$, we compute the final logits as
\begin{equation} 
h_i^{\prime}(\bm{s})_j=\sum_{j \leq l} \log p(a_{ij} \mid \bm{s}) \\
+\sum_{j>l} \log \left(1- p(a_{ij} \mid \bm{s}) \right),
\label{logits}
\end{equation}
and derive the final output probability using a Softmax $\hat{p}(a_{ij}) = Softmax(h_i^{\prime}(\bm{s})_j)$.
The implementation of this approach is illustrated in terms of the layers at the end of the deep network, as depicted in Fig.~\ref{frame}.

It is noteworthy that utilizing independent unimodal probability distributions for producing units guarantees the monotonic relationship ($\hat{p}_{i0}<\hat{p}_{i1}<\hat{p}_{i2},...,<\hat{p}_{ik}$ and $\hat{p}_{ik}>\hat{p}_{ik+1},...,>\hat{p}_{iK}$), which is desirable for imposing inequality constraints on the outputs to maintain monotonicity.
Combining unimodal probability distributions and the ordinal architecture, the unimodal policy makes the most sense as it allows for easy separation of samples drawn from action $k$ and those drawn from all other actions, enhancing the selected probability of the most conﬁdent action.
With such unimodal parameterization, the form of unimodal probability distributions can easily find maximal concerning actions while retaining a smaller space of neural architectures.

\subsection{Variance Analysis}
While PG algorithms are well-suited for parametric distribution functions, it is susceptible to converging to a poor policy that is not even a local optimum. 
One of the primary limitations of the plain policy gradient method is the high variance of the policy gradient estimator.
Using the Gaussian distribution as a stochastic policy in continuous control, $\pi_{\theta}(a\mid s)=\frac{1}{\sqrt{2 \pi} \sigma} \exp \left(-\frac{(a-\mu)^{2}}{2 \sigma^{2}}\right)$, has been well-studied and commonly used in the RL community since~\cite{williams1992simple}, because the Gaussian distribution is easy to sample and has gradients that are easy to compute.
However, one undesirable property of Gaussian policy is that the variance of the policy gradient estimator tends to increase as the standard deviation $\sigma^2$ of the Gaussian distribution decreases~\cite{chou2017improving}.
Generally, the policy gradient estimator is given as 
\begin{equation}
\hat{g}_{\theta}=r\left(a_{j}\right) \nabla_{\theta} \log \hat{p}_{\theta}(j), j \sim \hat{p}_{\theta}(j),
\label{estimator}
\end{equation}
and the variance can be represented as
\begin{equation}
\mathbb{V}\left[\hat{g}_{\theta}\right]=\mathbb{E}\left[\hat{g}_{\theta}^{2}\right]-\mathbb{E}^{2}\left[\hat{g}_{\theta}\right]
\label{variance}
\end{equation}
For Gaussian distribution, as the policy improves and becomes more deterministic ($\sigma \rightarrow 0$), the variance of the policy gradient estimator goes to inﬁnity, which makes PG methods sample-inefficient with respect to interactions with the environment and hinders the applications to real physical control systems.

As mentioned earlier, the variance of the Poisson distribution is equivalent to its mean, whose properties can alleviate the above high variance problem.
Before showing the theoretical analysis for the variance of unimodal policy, we first make the following assumption that considering a simplified setting of a one-step bandit problem with one dimension action space $\mathcal{A} = [-1,1]$ and fixed constant reward $r(a) = R$, for all actions $a$.
We denote $p(j)$ to be the probability of taking the $j$th action $0 \leq j \leq {K-1}$, and upon initialization, the policy has very high entropy $p{(j)} \approx \frac{1}{K}$.

Recall that the policy gradient estimator is given by
\begin{equation}
\hat{g}_{\theta}=r\left(a_{ij}\right) \nabla_{\theta} \log p_{\theta}(j), j \sim p_{\theta}(j).
\end{equation}
With such a setting, we can derive the expectation of the policy gradient estimator $\mathbb{E}\left[\hat{g}_{\theta}\right]=\nabla_{\theta} J\left(\pi_{\theta}\right)=0$ and the variance is 
\begin{equation}
\begin{aligned}
\mathbb{V}\left[\hat{g}_{\theta}\right] &= \sum_{j=0}^{K-1} r^{2} \left(a_{j}\right) \left(\nabla_{\theta} \log p(j)\right)^{2} p(j) \\
& \approx R^{2} \frac{1}{K} \sum_{j=0}^{K-1}\left(\nabla_{\theta} \log p(j)\right)^{2} \\
& \approx R^{2} \frac{1}{K} \sum_{j=0}^{K-1} \left(\nabla_{\theta} L_{j}-\frac{1}{K} \sum_{k=0}^{K-1} \nabla_{\theta} L_{k}\right)^{2},
\end{aligned}
\label{equ14}
\end{equation}
where $\nabla_{\theta} L_{j}$ denotes the $j$th logit independently depending on $\theta$, and the approximations come from $p{(j)} \approx \frac{1}{K}$.
In the context of standard neural network initialization schemes, wherein all weight and bias matrices (e.g., $\theta$, $w_{j}^T h$, and $b_{j}$) are initialized independently, the gradients $\nabla_{\theta} L_{j}$ can be regarded as identically and independently distributed (i.i.d.) random variables. 
The stochastic nature of these variables originates from the random initialization of the neural network parameters.
We employ the notation $\mathbb{E}_{\text{init}}[\cdot]$ to represent the expectation concerning neural network initializations. 
With this notation in place, we analyze the expectation of the specified equation of (\ref{equ14}) as follows:
\begin{equation}
\begin{aligned}
\mathbb{E}_{\text {init }}\left[\mathbb{V}\left[\hat{g}_{\theta}\right]\right] 
&\! \approx \frac{R^{2}}{K} \mathbb{E} \left(  \sum_{j=0}^{K-1} \left( \nabla_{\theta} L_{j} - \frac{1}{K} \sum_{k=0}^{K-1} \nabla_{\theta} L_{k} \right)^{2} \right) \\
&\! = \frac{R^{2}}{K} \mathbb{E} \left(\sum_{i=1}^{N}\sum_{j=0}^{K-1} \left( \nabla_{\theta} L_{j} - {\mu} + {\mu} - \mathbb{E} \left(\nabla_{\theta} L\right)  \right)^{2}\right) \\
&\! = \frac{R^{2}}{K}  \sum_{j=1}^{K} \mathbb{E}\left( \nabla_{\theta} L_{j} - {\mu} \right)^{2} - K (\mathbb{E} \left(\nabla_{\theta} L\right)-\mu)^{2} \\
&\! = \frac{R^{2}}{K}  \left( K \mathbb{V}_{\text {init}} \left( \nabla_{\theta} L\right) - K \mathbb{V}_{\text {init}} (\mathbb{E} \left(\nabla_{\theta} L\right)) \right) \\
&\! = \frac{R^{2}}{N}   \left( \mathbb{V}_{\text {init}} \left( \nabla_{\theta} L\right) - \mathbb{V}_{\text {init}} (\mathbb{E} \left(\nabla_{\theta} L\right)) \right),
\end{aligned}
\label{equ15}
\end{equation}
where $\mu$ is the exception of $\nabla_{\theta} L$.

Let $\hat{g}_{\theta_1}$ and $\hat{g}_{\theta_2}$ denote the policy gradient estimator using the Poisson distribution policy and the ordinal distribution policy, respectively.
By analyzing the expectation of~(\ref{equ15}), we can derive the expressions for the variances of the policy gradient estimators of the Poisson distribution
\begin{equation}
\begin{aligned}
\mathbb{E}_{\text {init }}\left[\mathbb{V}\left[\hat{g}_{\theta_1}\right]\right] 
& = R^{2}  \left( \mathbb{V}_{\text {init}} \left( \nabla_{\theta_1} L\right) - \mathbb{V}_{\text {init}} (\mathbb{E} \left(\nabla_{\theta_1} L\right)) \right) \\
& \approx R^{2} \frac{K-1}{K} \sigma_{1}^{2} \sim \frac{K-1}{K},
\label{equ16}
\end{aligned}
\end{equation}
where $\sigma_{1}^{2}=\mathbb{V}_{\text {init}}\left[\nabla_{\theta_1} L_{j}\right]$ is the variance of the PMF of Poisson distribution produced by one logit.
Then we analyze the expectation of~(\ref{equ15}) with ordinal probability distribution:
\begin{equation}
\mathbb{E}_{\text {init }}\left[\mathbb{V}\left[\hat{g}_{\theta_2}\right]\right] \approx R^{2} \frac{K-1}{K} \sigma_{2}^{2} \sim \frac{K-1}{K},
\end{equation}
where $\sigma_{2}^{2}=\mathbb{V}_{\text {init }}\left[\nabla_{\theta_2} L_{j}\right]$ is the variance of the $K$ logits gradients.
Considering the dependence of $\nabla_{\theta_2} L_{j}$ on $\theta_2$, it has been argued by~\cite{tang2020discretizing} that finer discretization (larger $K$) theoretically leads to increased variance for the $L_{j}$ of the neural network output, which may consequently exacerbate model variance and degrade performance. 
One key advantage of employing the Poisson distribution in statistical applications is the equivalence of variance and mean, suggesting that the variance is only mildly influenced by widespread probability mass distribution, particularly with a large action space.

In practice, we opt for larger $K$ values for refined action selection, harnessing the properties of the Poisson probability distribution as opposed to the ordinal policy, which exhibits greater variance and renders policy optimization more difficult, especially in the context of finer discretization. 
Generally, a moderate action discretization $K$ (e.g., $9\leq K \leq15$) strikes an appropriate balance between performance enhancement and computational expense.
Nevertheless, using the Poisson distribution presents a drawback when truncating infinite probability mass distributions. 
Inspired by~\cite{da2008unimodal}, the infinite support issue is tackled using a truncated Poisson distribution, normalizing probabilities to ensure that their summation equals one. 
This strategy closely mirrors our approach, which incorporates a Softmax operation. 
To regulate the distribution variance induced by the Softmax operation, a judicious selection of $\tau$ can help alleviate this concern. 
Our unimodal policy implementation spans a range of $\tau=[1.5, 3]$ (with varying $K \in [9,15]$).
As a validation of our approach, we assess the proposed methods on an array of complex control tasks, yielding substantial performance gains. 
These findings imply that our methodology benefits from the diminished variance of constrained neural network output units, which induce unimodal probability distributions characterized by low variance.

\section{Experiments}\label{Sec4}
The goal of our experimental evaluation is to investigate the improvement and stability of our method in comparison to baseline algorithms and previous Gibbs policy approaches on continuous control tasks.
To distinguish between the different policies, we will use the terms Gibbs policy for the Gibbs distribution, ordinal policy for the ordinal distribution, and unimodal policy for the unimodal architecture.
First, we evaluate our method against the Gaussian policy, employing baseline algorithms (TRPO~\cite{schulman2015trust}, PPO~\cite{schulman2017proximal}, and ACKTR~\cite{wu2017scalable}) on challenging continuous control tasks provided by the OpenAI gym benchmark suite~\cite{brockman2016openai}.
We consider Gaussian policy as a comparison with special attention since it is the default policy class implemented in on-policy baselines base~\cite{baselines}.
Additionally, we explore other architectural alternatives, such as Gaussian with tanh non-linearity in the output layer and Beta distribution, both of which have been proposed in prior works~\cite{chou2017improving, tang2020discretizing}.
Since a wide range of existing algorithms can solve the easier tasks, we pay more attention to the high dimension tasks, such as the 17-dimensional Humanoid~\cite{brockman2016openai}. 
Further, we compare the Gibbs policy and ordinal policy, which are prior representative techniques.
Finally, we study the stability of our algorithm, which plays a large role in performance and practicals.
We implement all the methods with Tersorflow 1.15.0 framework in Python 3.6 running on Ubuntu 18.04 with 2 AMD EPYC 7H12 64-Core CPU Processors and 3 NVIDIA GeForce RTX 3080 GPUs.

\textbf{Implementation Details.}
Our investigation focuses on the impact of the discrete unimodal probability distribution policy in conjunction with on-policy RL algorithms. 
To this end, we introduce minimal alterations to the original TRPO, PPO, and ACKTR algorithms, as implemented in OpenAI baselines~\cite{chou2017improving}. The implementations for all algorithms (PPO, TRPO, ACKTR) are grounded in OpenAI baselines. 
We conduct a comparative analysis of our method against previous approaches on a diverse range of challenging continuous control tasks, as provided by the OpenAI gym benchmark suite, with action constraints set within the $[-1,1]$ interval as a normalizing flow. 
Detailed descriptions of each policy class utilized in our experiments are provided below.
The implementation code can be found in~\url{https://anonymous.4open.science/r/udprl-tnnls}.

\textbf{\emph{Gaussian Policy.}}
The factorized Gaussian policy~\cite{heess2015learning,schulman2015trust,schulman2017proximal} is represented as $\pi_{\theta}(\cdot \mid s)=\mathcal{N}\left(\mu_{\theta}(s), \sigma^2\right)$, where the mean $\mu_{\theta}(s)$ is derived from a two-layer neural network consisting of $128$ units per layer for PPO and ACKTR, and $128$ units per layer for TRPO.
The diagonal standard deviation $\sigma^2$ is defined as $\sigma^{2}_{ii}= \sigma^{2}_i$, with each $\sigma_i$ being a single variable shared across all states.
We adhere to the default hyperparameter settings established in the baselines.

\textbf{\emph{Gibbs Policy.}}
The Gibbs policy~\cite{tang2020discretizing} is represented as $\pi_{\theta}(\cdot \mid s)=\frac{\exp \left(h_{i}(s \mid \boldsymbol{\theta}) \right)/\tau}{\sum_{j \in A(s)} \exp \left(h_{j}(s \mid \boldsymbol{\theta})\right)/\tau}$ with Gibbs distribution over the discrete actions space, which is parameterized by a neural network that shares the same architecture as the Gaussian policy described above.
The temperature parameter $\tau$ is set to $1.5$.
We define $K$ evenly spaced atomic actions for every dimension, ranging from $-1$ to $1$.
For each dimension of the action space, the discrete distribution encompasses $K$ logits $L_j(s)$, where $L_j(s)$ denotes the logits distribution of the $j$-th action in state $s$.

\textbf{\emph{Ordinal Policy.}}
The ordinal policy~\cite{tang2020discretizing} differentiates it from discrete policies by incorporating an ordinal parameterization while maintaining the same number of parameters.

\textbf{\emph{Unimodal Policy.}}
In contrast to ordinal policies, unimodal policies utilize a Poisson probability distribution to enforce unimodality. 
 The scalar output of our deep network for unimodal policies is represented as $[f_1(\bm{s}), f_2(\bm{s}),..., f_i(\bm{s})]$ rather than the $K$ logits represented in discrete policies, where $f_i(\bm{s})$ indicates the average frequency of events for the $i$-dimension atomic actions. 
 The policy parameters are consistent with ordinal policies, except for the addition of a Softplus nonlinearity transformation in the final layer to ensure $f_i(\bm{s}) >0$. 
 The temperature parameter $\tau$ is set to $2.5$ for varying discretization $K={9,11,15}$.

\textbf{\emph{Gaussian+tanh Policy.}}
The Gaussian-tanh policies~\cite{haarnoja2018soft,tang2020discretizing} are the same as the Gaussian policies above, but the final layer is added a tanh transformation to ensure the mean $\mu_{\theta} \in [-1, 1]$.

\textbf{\emph{Beta Policy.}}
The Beta policy~\cite{chou2017improving} are represented as $\pi_{\theta}(\cdot \mid s)=f\left(\frac{a+h}{2 h} \mid \alpha, \beta\right)$, where $h$ is max of the closed interal of the action space $A=[-h, h]$. 
Parameters $\alpha=\alpha_{\theta}(s)$ and $\beta=\beta_{\theta}(s)$ are determined by a two-layer neural network $f_{\theta}(s)$, incorporating a Softplus activation function in its final layer.
It is important to note that this parameterization can cause instability throughout the learning process, particularly when $\alpha_{\theta}(s)\rightarrow \infty$ or $\beta_{\theta}(s) \rightarrow \infty$, leading to unstable dynamics. 
In practice, this instability is apparent when the trust region size is large (e.g., $\epsilon=0.005$ and $\epsilon=0.01$), and can result in training termination due to numerical errors. 
Conversely, a smaller trust region size (e.g., $\epsilon=0.001$) may yield diminished performance. 
In our implementation, we employ the main thread settings for the Beta policy, which are derived with a trust region size of $\epsilon=0.01$ for TRPO.

\textbf{\emph{Others Hyperparameters.}}
For value functions, two-layer neural networks are utilized, consisting of $128$ hidden units per layer in the case of PPO and ACKTR, and $64$ hidden units per layer for TRPO. 
We used the Adam optimizer (Kingma and Ba 2015) with a learning rate $2 \cdot 10^{-5}$, $\beta_{1}=0.9$, and $\beta_{2}=0.999$ for both PPO and TRPO.
We trained with a minibatch size of 64 and a discount factor $\gamma=0.98$.
The KL constraint parameter, $\epsilon$, is set within the range of ${0.01, 0.001}$ for TRPO and ${0.02, 0.002}$ for ACKTR. 
All remaining hyperparameters align with those found in the original baseline implementations.
\begin{figure*}[ht]
\begin{center}
\subfigure[Hopper-v3 + PPO]{\includegraphics[height=3.31cm]{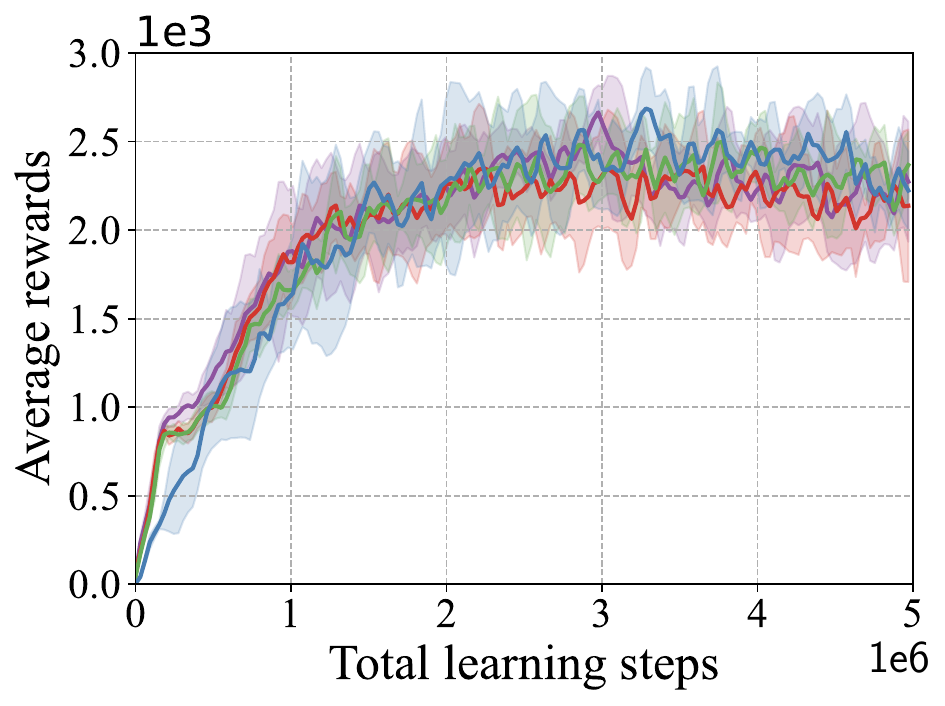}}
\subfigure[Walker2D-v3 + PPO]{\includegraphics[height=3.31cm]{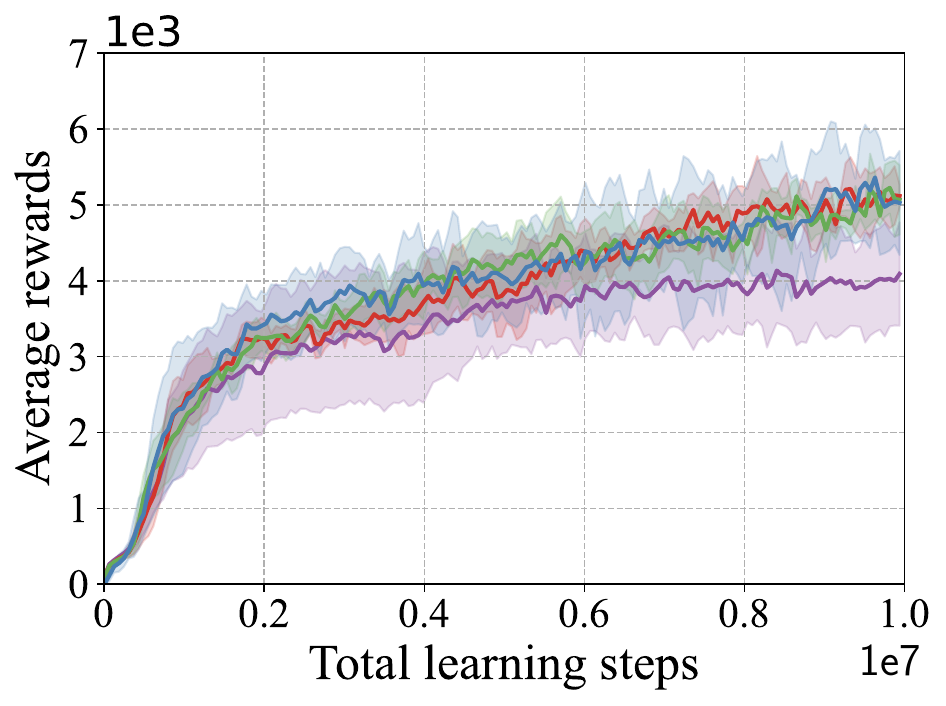}}
\subfigure[Ant-v3+ PPO]{\includegraphics[height=3.31cm]{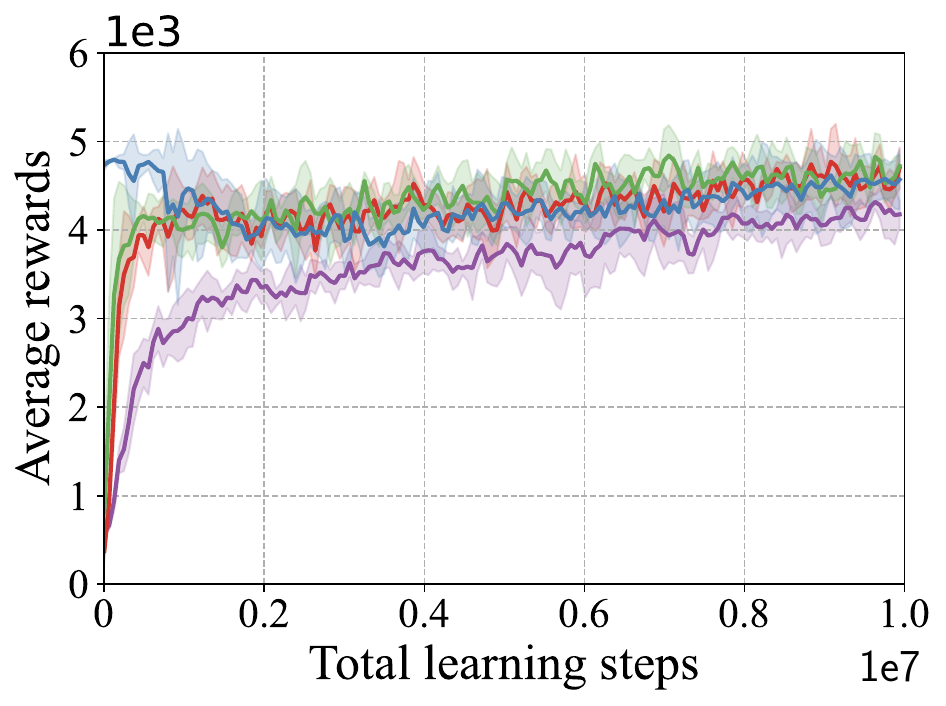}}
\subfigure[HalfCheetah-v2+ PPO]{\includegraphics[height=3.31cm]{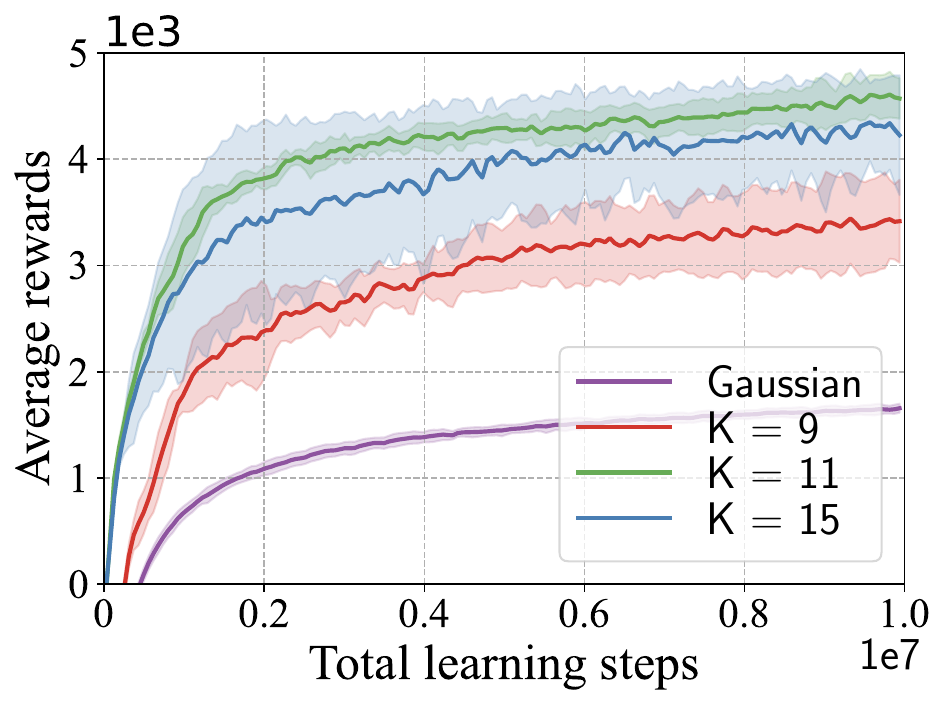}}
\subfigure[HalfCheetah-v3+ PPO]{\includegraphics[height=3.31cm]{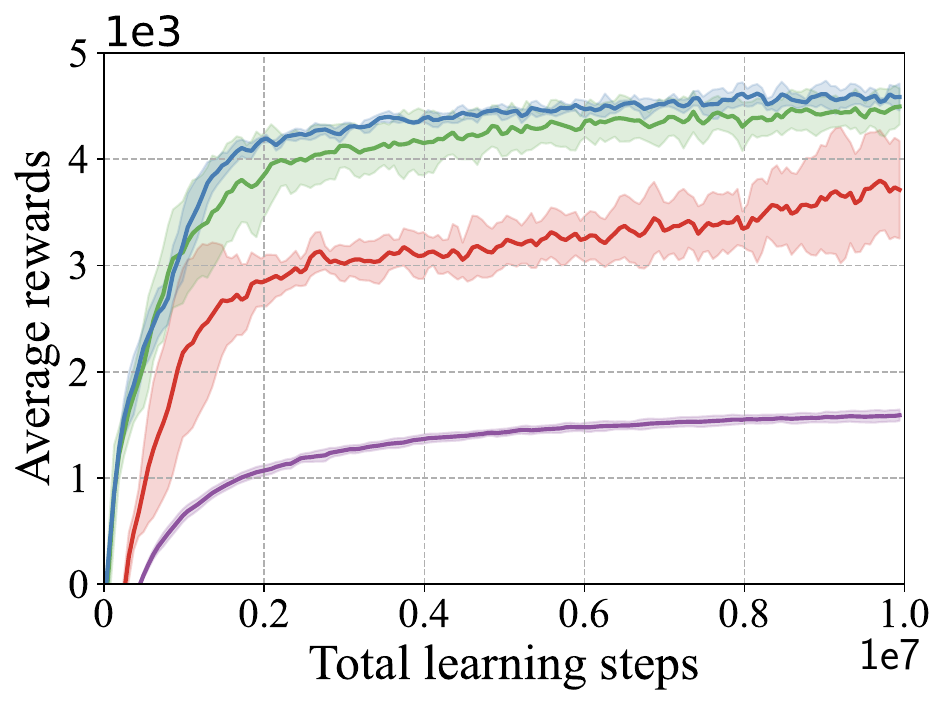}}
\subfigure[Humanoid-v2+ PPO]{\includegraphics[height=3.31cm]{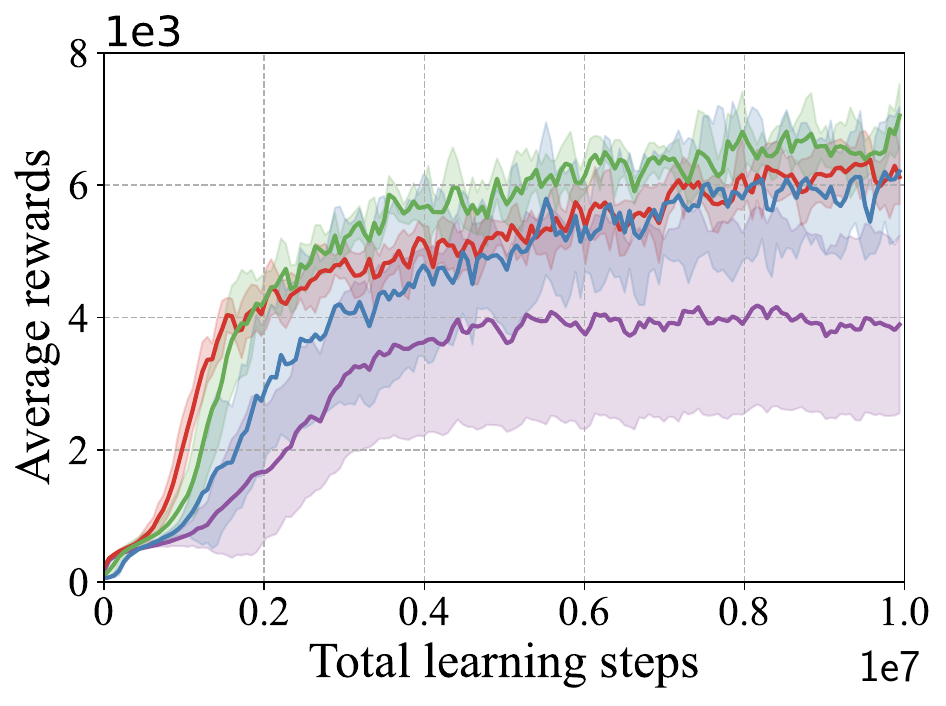}}
\subfigure[Humanoid-v3+ PPO]{\includegraphics[height=3.31cm]{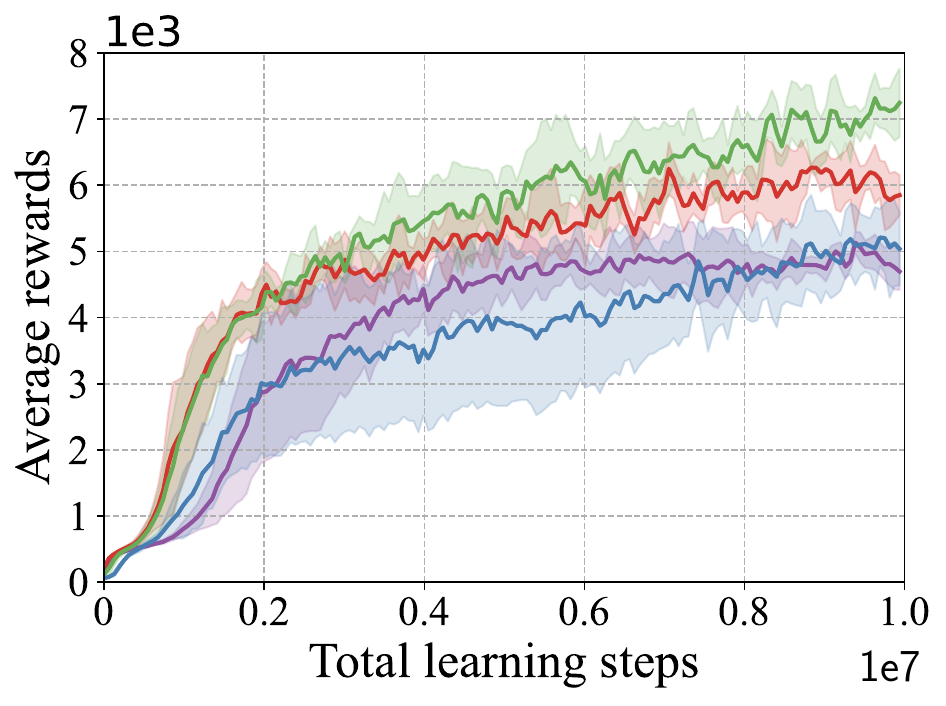}}
\subfigure[HumanoidStandup-v2+ PPO]{\includegraphics[height=3.31cm]{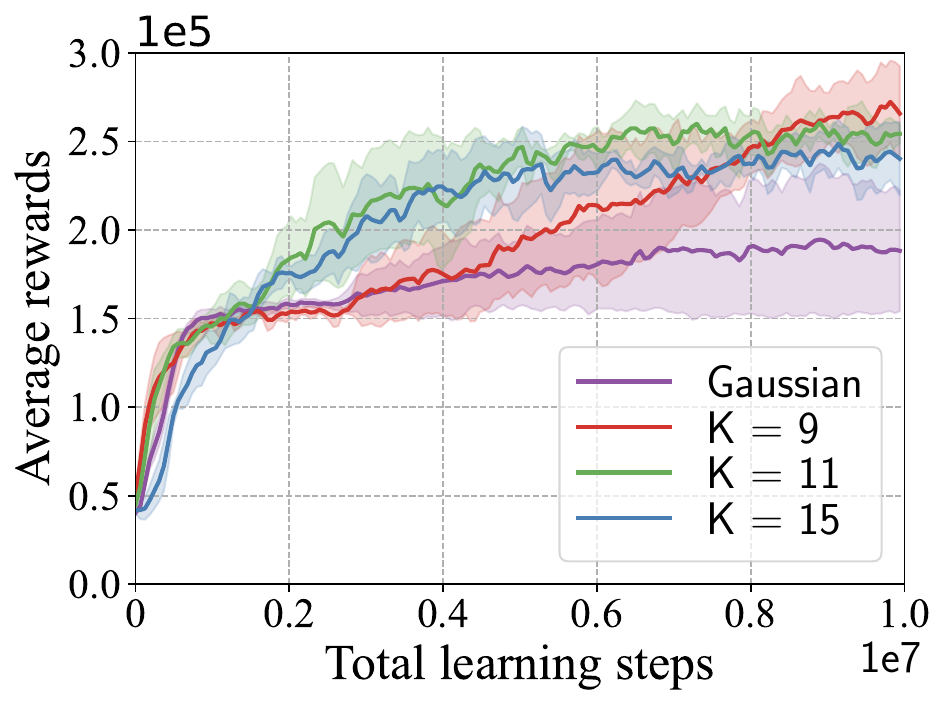}}
\subfigure[Hopper-v3 + TRPO]{\includegraphics[height=3.31cm]{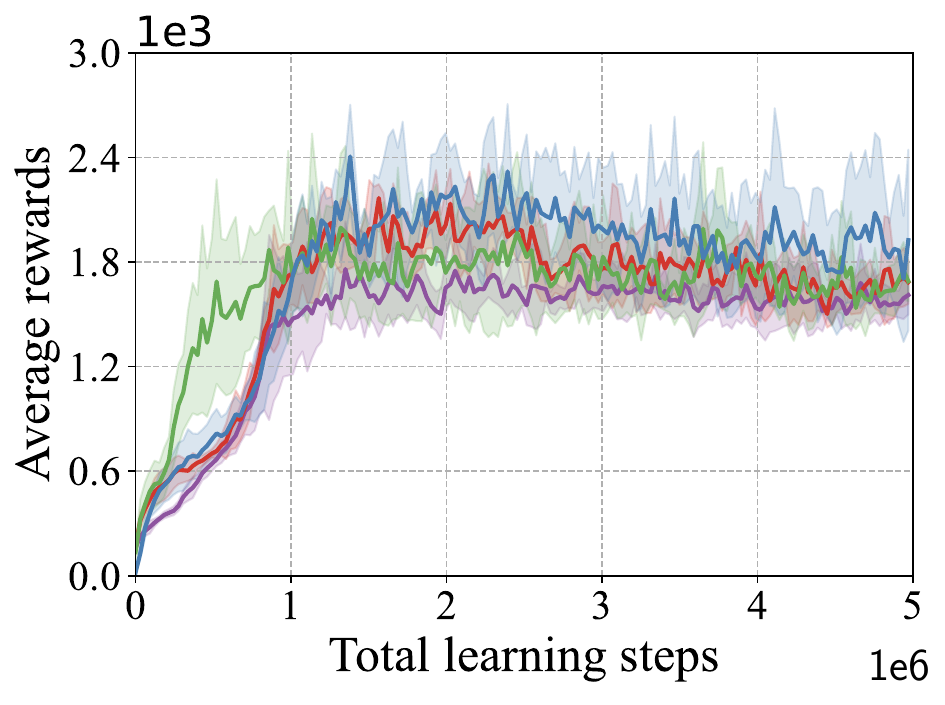}}
\subfigure[Walker2D-v3 + TRPO]{\includegraphics[height=3.31cm]{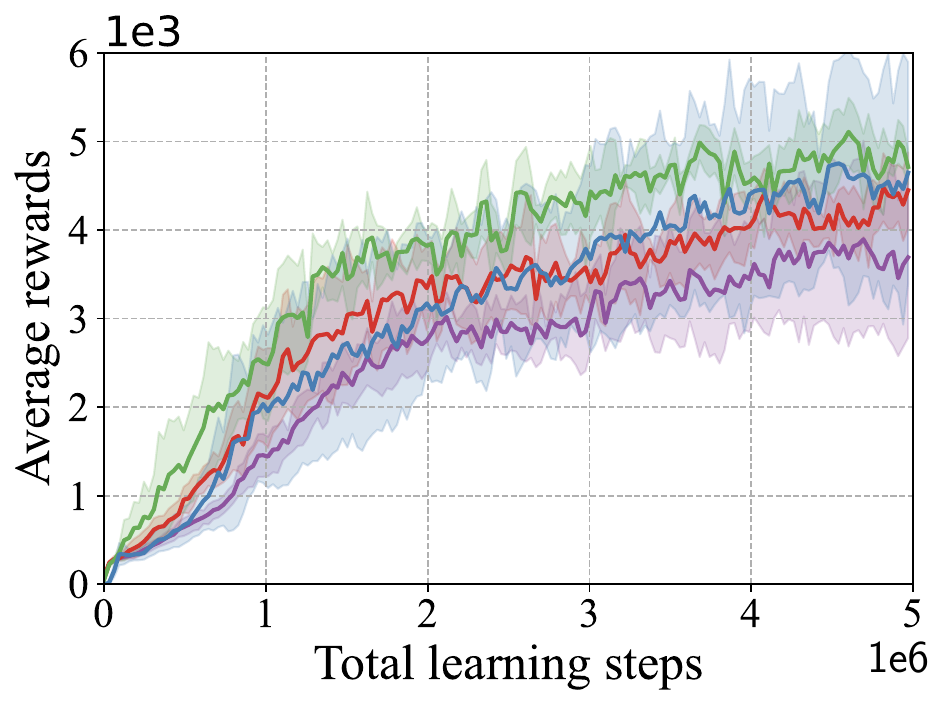}}
\subfigure[Ant-v3+ TRPO]{\includegraphics[height=3.31cm]{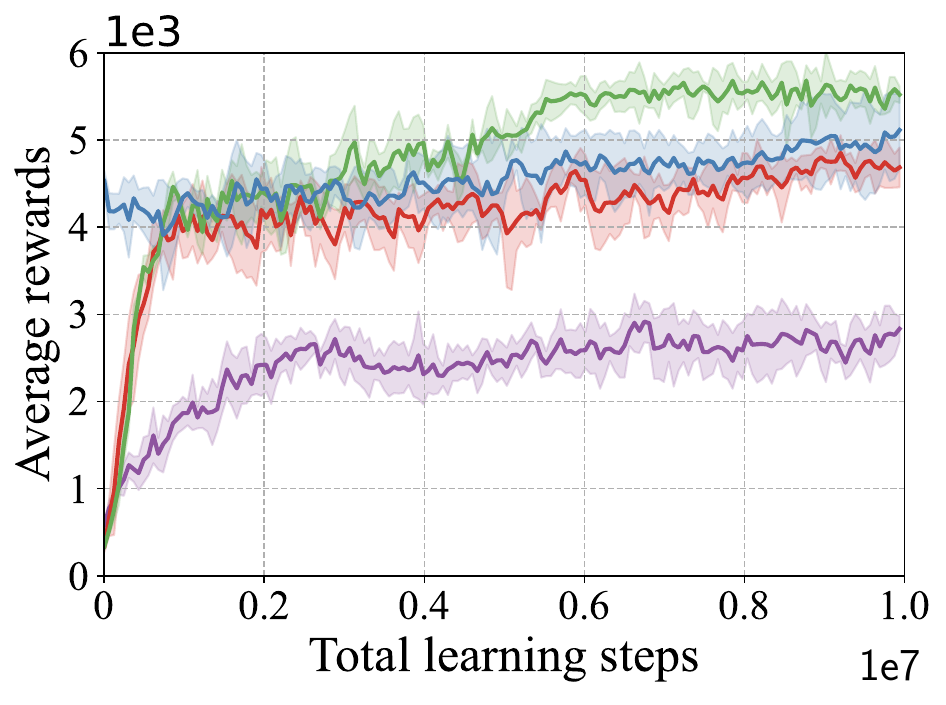}}
\subfigure[HalfCheetah-v2+ TRPO]{\includegraphics[height=3.31cm]{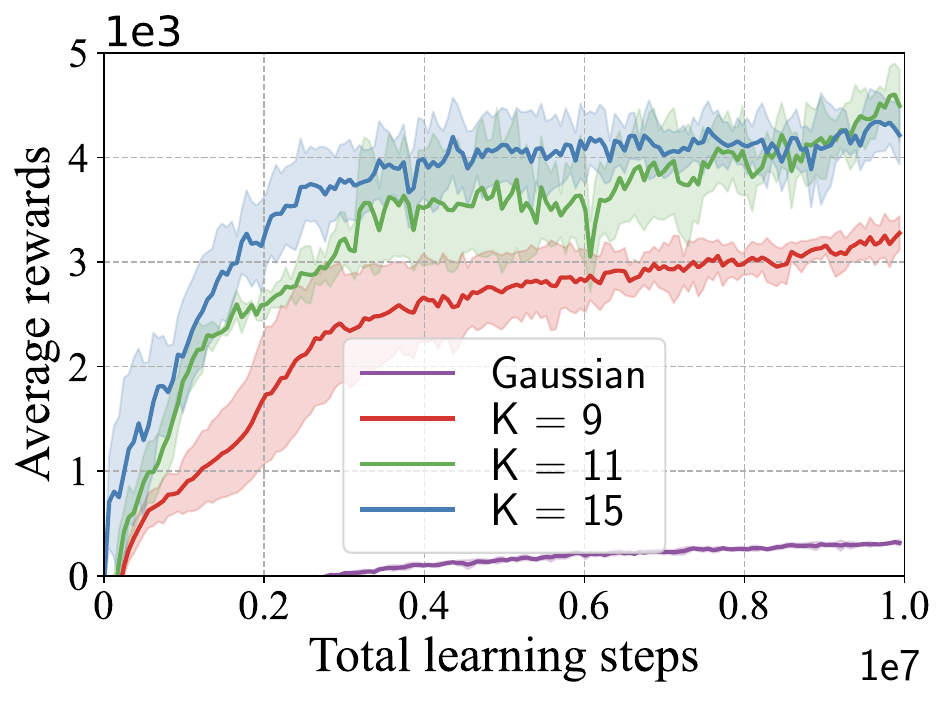}}
\subfigure[HalfCheetah-v3+ TRPO]{\includegraphics[height=3.31cm]{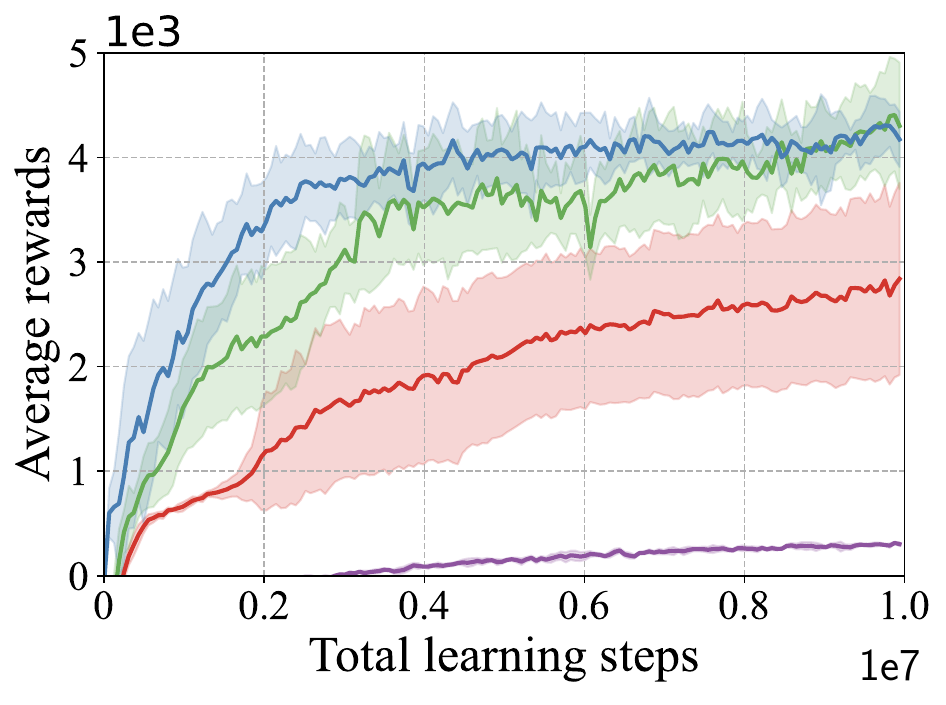}}
\subfigure[Humanoid-v2+ TRPO]{\includegraphics[height=3.31cm]{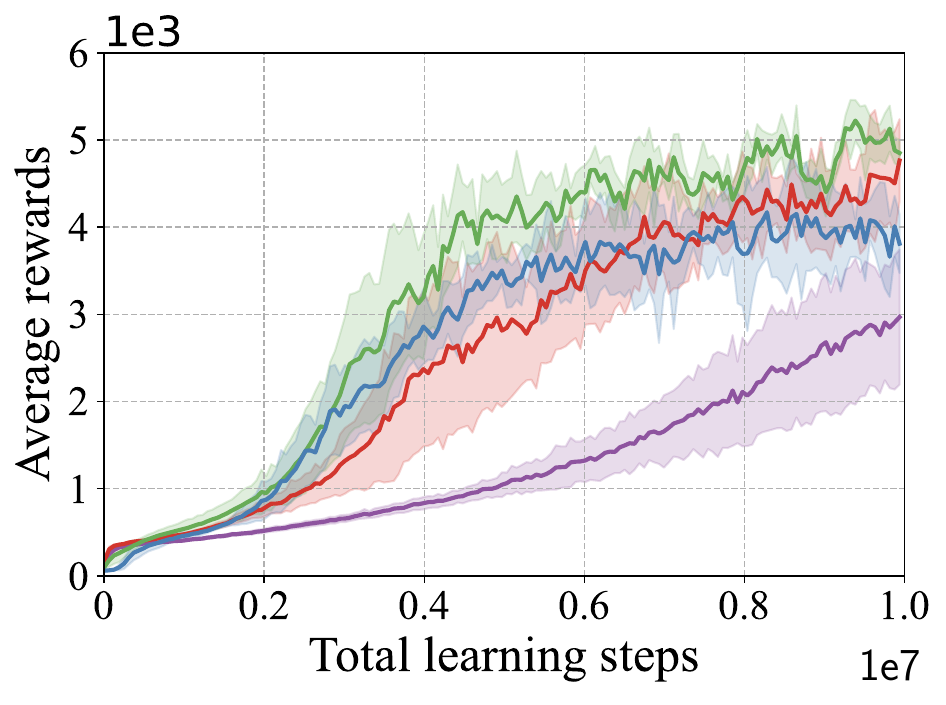}}
\subfigure[Humanoid-v3+ TRPO]{\includegraphics[height=3.31cm]{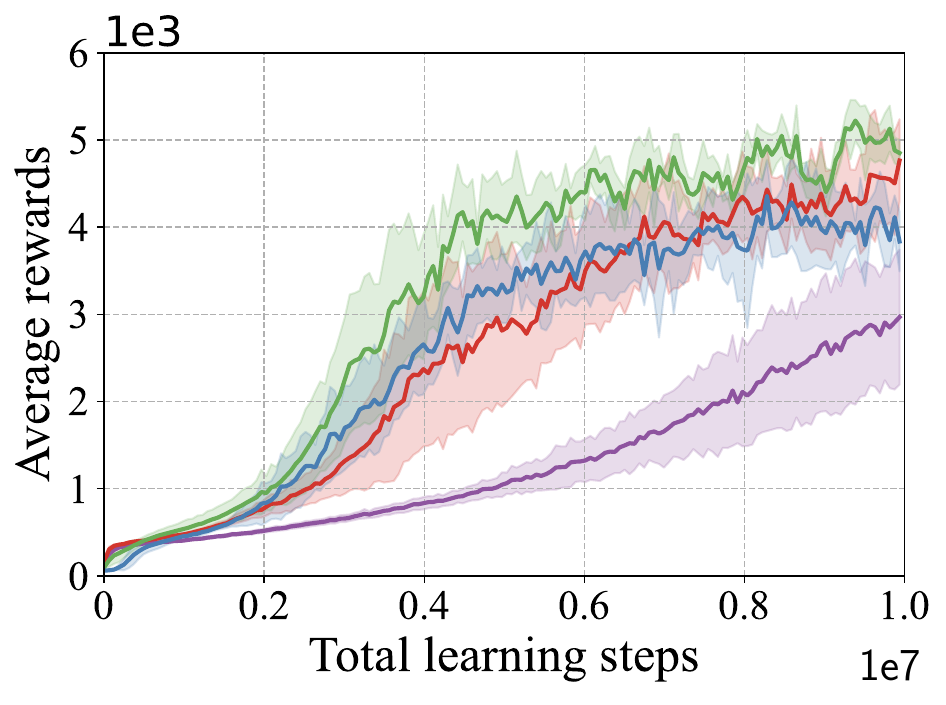}}
\subfigure[HumanoidStandup-v2+ TRPO]{\includegraphics[height=3.31cm]{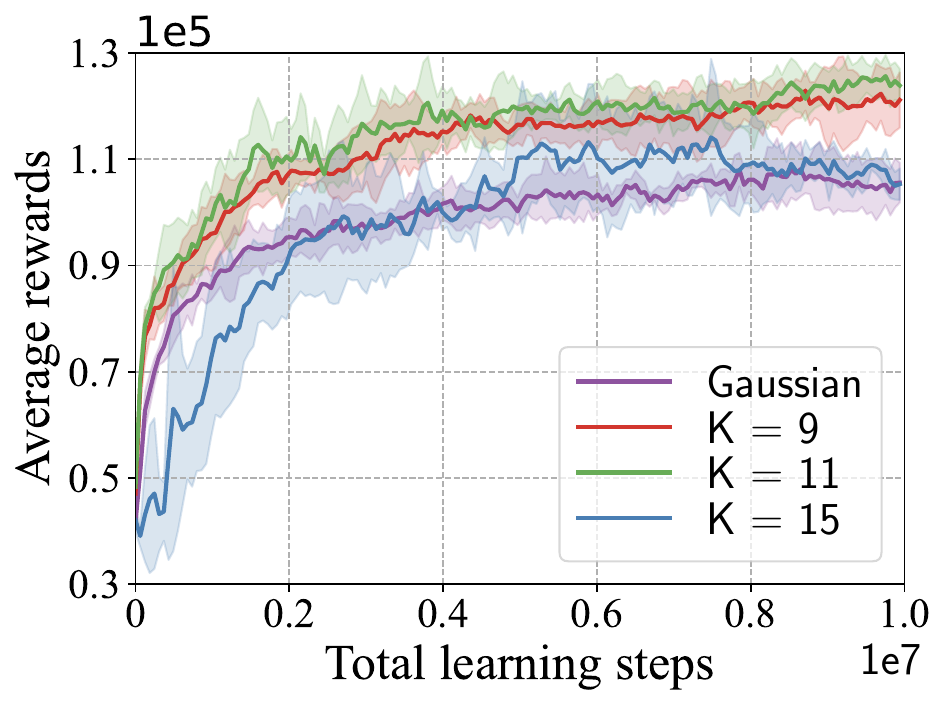}}
\caption{Performance as a function of the number of learning steps of PPO and TRPO on OpenAI gym MuJoCo locomotion tasks.
Solid lines are average values over $6$ random seeds.
Shaded regions correspond to one standard deviation.
Each curve corresponds to a different policy architecture (Gaussian or unimodal policy with varying bins $K = 9, 11,15$).
Our unimodal policy significantly outperforms the Gaussian policy on most tasks.}
\vskip -0.2in
\label{exp1}
\end{center}
\end{figure*}
\begin{table*}[tb]
\caption{Comparison against the representative baseline policy (Gaussian, Gaussian + tanh, Beta~\cite{chou2017improving}, Gibbs and ordinal policy distribution) implemented in PPO and TRPO. Average $\pm$ std return over the last $20$ evaluations over $10$ trials of $10$ million time steps. The result with the top two highest cumulative rewards for each task is bolded.}
\begin{center}
\scalebox{1.01}{
\begin{tabular}{lcccccc}
\hline \textbf{PPO} & \textbf{Gaussian} & \textbf{Gaussian+tanh} & \textbf{Beta} & \textbf{Gibbs} & \textbf{Ordinal} & \textbf{Unimodal (ours)} \\
\hline Hopper & $\mathbf{1962} \pm \mathbf{313}$	& $1781 \pm 295$	 & $964 \pm 1003$	 & $1696 \pm 140$	& $1808 \pm 169$ & $\mathbf{1858} \pm \mathbf{230}$\\
Walker2D & $3991 \pm 676$	& $3560 \pm 486$	& $179 \pm 25$	& $\mathbf{4572} \pm \mathbf{269}$		& $3994 \pm 291$ & $\mathbf{4948} \pm \mathbf{295}$\\
Ant-v3 & $4193 \pm 201$	& $4413 \pm 247$	& $\mathbf{5356} \pm \mathbf{540}$	& $\mathbf{6033} \pm \mathbf{268}$		& $5154 \pm 368$ & $4650 \pm 257$\\
Halfcheetah-v2 & $1645 \pm 44$	& $1584 \pm 33$	& $906 \pm 22$	& $\mathbf{5052} \pm \mathbf{116}$		& $1783 \pm 912$ & $\mathbf{4569} \pm \mathbf{191}$\\
Halfcheetah-v3	& $\mathbf{1586} \pm \mathbf{53}$ 	& $269 \pm 25$ 		& $381 \pm 79$ 		& $313 \pm 24$		& $1525 \pm 151$ & $\mathbf{4586} \pm \mathbf{97}$\\
Humanoid-v1		& $3895 \pm 1368$ 	& $5070 \pm 414$ 	& $1200 \pm 1974$	& $\mathbf{5327} \pm \mathbf{422}$  & $4970 \pm 384$ & $\mathbf{6593} \pm \mathbf{378}$\\
Humanoid-v3		& $4886 \pm 226$ 		& $4927 \pm 280$ & $791 \pm 36$  & $\mathbf{5443} \pm \mathbf{280}$& $4941 \pm 204$ & $\mathbf{7072} \pm \mathbf{409}$\\
HumanoidStandup-v2 & $189911 \pm 36927$ 	& $\mathbf{220065} \pm \mathbf{49295}$	& $34470 \pm 9507$ 	& $159043 \pm 2660$		& $158247 \pm 2402$ & $\mathbf{266878} \pm \mathbf{24353}$\\

\hline \textbf{TRPO} & & & &  \\
\hline Hopper & $1588 \pm 119$ 	& $1125 \pm 114$ 	& $1421 \pm 22$ 		& $\mathbf{1685} \pm \mathbf{140}$		& $1645 \pm 112$ & $\mathbf{1725} \pm \mathbf{197}$\\
Walker2D & $3779 \pm 921$ 		& $1344 \pm 127$ 		& $980 \pm 231$  		& $\mathbf{3974} \pm \mathbf{341}$		& $3866 \pm 289$ & $\mathbf{4918} \pm \mathbf{319}$\\
Ant-v3 & $2674 \pm 225$ 		& $2844 \pm 229$ 		& $4317 \pm 576$  		& $\mathbf{5186} \pm \mathbf{176}$		& $4749 \pm 130$ & $\mathbf{5556} \pm \mathbf{187}$\\
Halfcheetah-v2 & $303 \pm 20$ 		& $319 \pm 367$ 		& $1392 \pm 278$  		& $\mathbf{1419} \pm \mathbf{16}$		& $1362 \pm 39$ & $\mathbf{4415} \pm \mathbf{236}$\\
Halfcheetah-v3			& $295 \pm 21$ 	& $341 \pm 26$ 		& $963 \pm 78$ 		& $2730 \pm 865$		& $\mathbf{4184} \pm \mathbf{250}$ & $\mathbf{4252} \pm \mathbf{465}$\\
Humanoid-v2	& $2837 \pm 677$ 	& $805 \pm 25$ 	& $512 \pm 59$	& $4646 \pm 321$	 & $\mathbf{4661} \pm \mathbf{139}$ & $\mathbf{4993} \pm \mathbf{242}$\\
Humanoid-v3		& $2955 \pm 722$ 		& $834 \pm 47$ 		& $485 \pm 31$  		& $\mathbf{4464} \pm \mathbf{480}$		& $4041 \pm 373$ & $\mathbf{4993} \pm \mathbf{242}$\\
HumanoidStandup-v2& $105163 \pm 4942$ 	& $111572 \pm 7139$ 		& $81724 \pm 6264$ 	& $106971 \pm 220$		& $\mathbf{112851} \pm \mathbf{10533}$ & $\mathbf{124590} \pm \mathbf{3794}$\\
\hline
\label{texp1}
\end{tabular}}
\end{center}
\vskip -0.2in
\end{table*}
\begin{table*}[tb]
\caption{The table presents the performance of PPO + ordinal/unimodal/Gaussian policy against state-of-the-art baseline algorithms over $6$ random seeds on Humanoid series benchmark tasks from OpenAI gym. Average return over the last 20 evaluations over $10$ trials of $10$ million time steps. The outcomes for DDPG, SQL, SAC, TD3, and DecQN are estimated based on the figures provided by~\cite{haarnoja2018soft}, and the PPO results align with the findings in the same study. The result with the highest cumulative rewards for each task is bolded.}
\begin{center}
\scalebox{0.91}{
\begin{tabular}{lcccccccc}
\hline \textbf{Tasks} & \textbf{DDPG} & \textbf{SQL} & \textbf{SAC} & \textbf{TD3} &\textbf{DecQN} & \textbf{PPO+Gaussian} & \textbf{PPO+ordinal}& \textbf{PPO+unidomal (ours)}\\
\hline Humanoid-v1 & $<500$  & $\approx 5500$  & $\approx 6000$  & $\approx 6000$ & $6271 \pm 904$ & $3895 \pm 1368$  & $4970 \pm 384$ & $\mathbf{6593} \pm \mathbf{378}$\\
Humanoid-v3 & $<500$  & $\approx 5500$  & $\approx 6000$  & $\approx 2000$ & $6546 \pm 658$ & $4886 \pm 226$  & $4941 \pm 204$ & $\mathbf{7072} \pm \mathbf{409}$\\
HumanoidStandup-v2 & $\approx 150000$ 	& $\approx 190000$	& $\approx 220000$ & $\approx 130000$ & $238439 \pm 25726$	& $189911 \pm 36927$ & $158247 \pm 2402$ & $\mathbf{266878} \pm \mathbf{24353}$\\
\hline
\label{texp2}
\end{tabular}}
\end{center}
\vskip -0.3in
\end{table*}
\begin{figure*}[ht]
\begin{center}

\subfigure[Hopper-v3 + PPO]{\includegraphics[height=3.31cm]{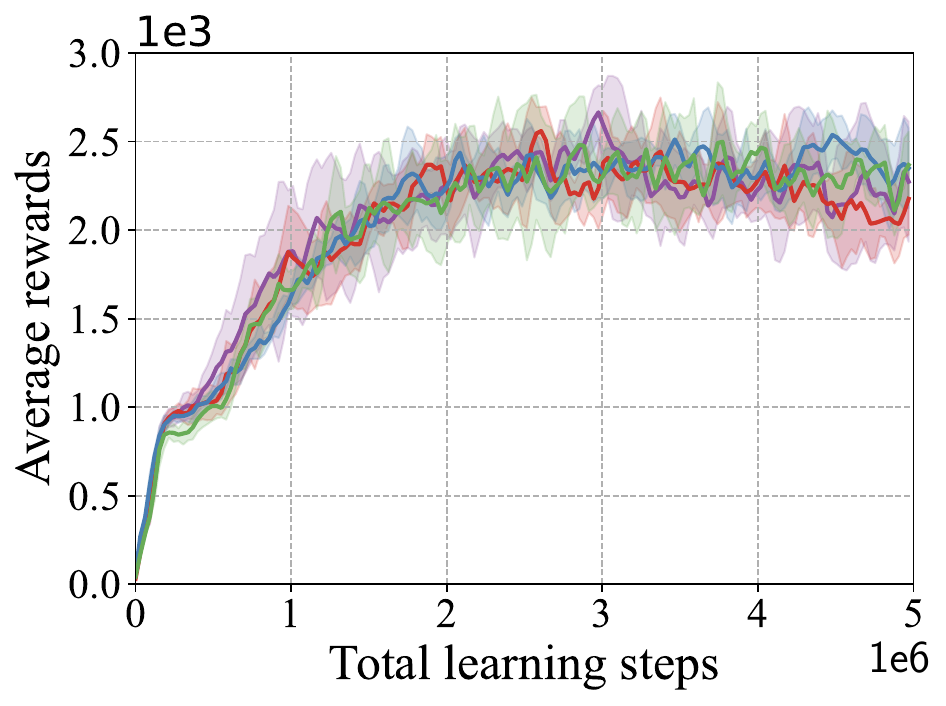}}
\subfigure[Walker2D-v3 + PPO]{\includegraphics[height=3.31cm]{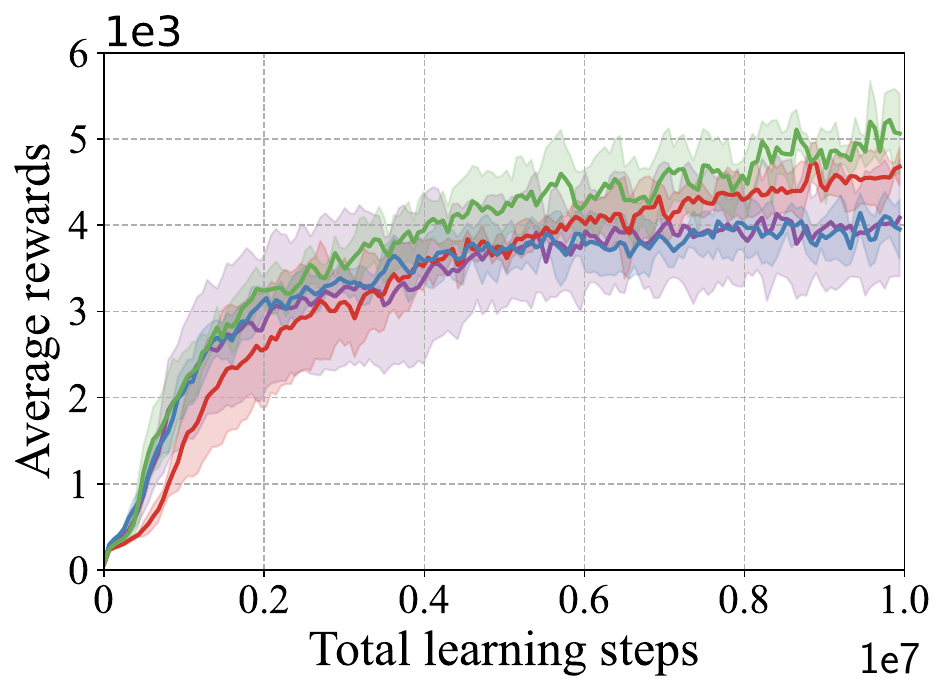}}
\subfigure[Ant-v3 + PPO]{\includegraphics[height=3.31cm]{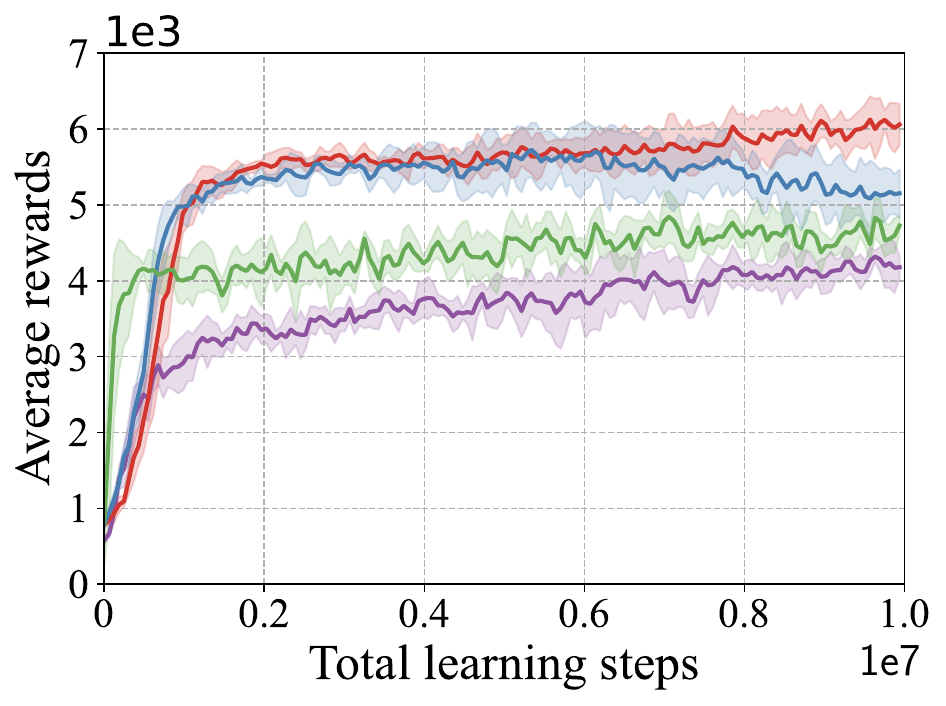}}
\subfigure[HalfCheetah-v2 + PPO]{\includegraphics[height=3.31cm]{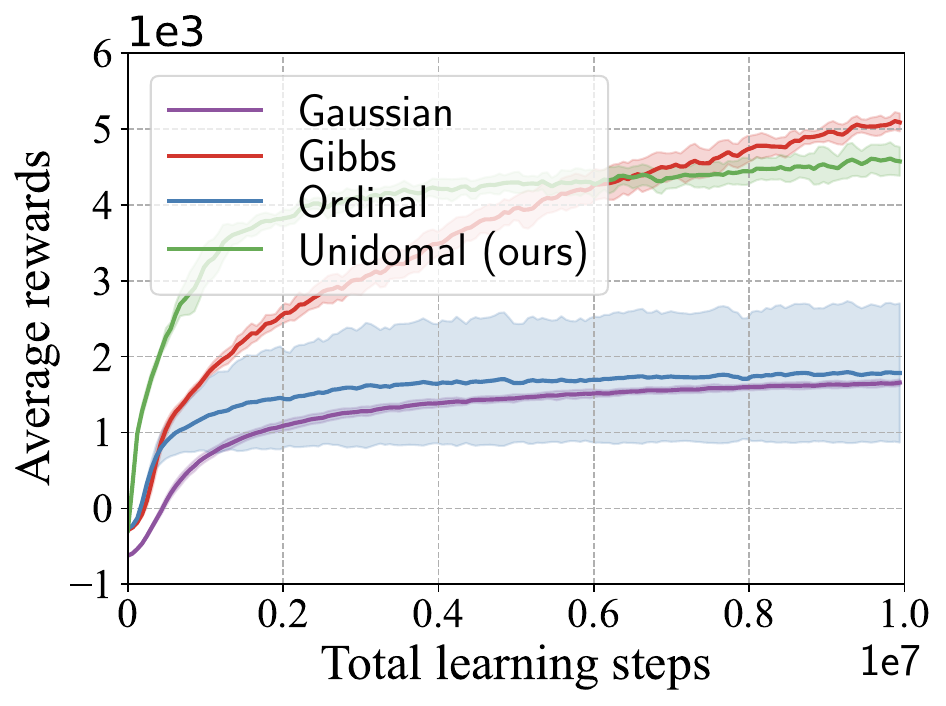}}
\subfigure[HalfCheetah-v3 + PPO]{\includegraphics[height=3.31cm]{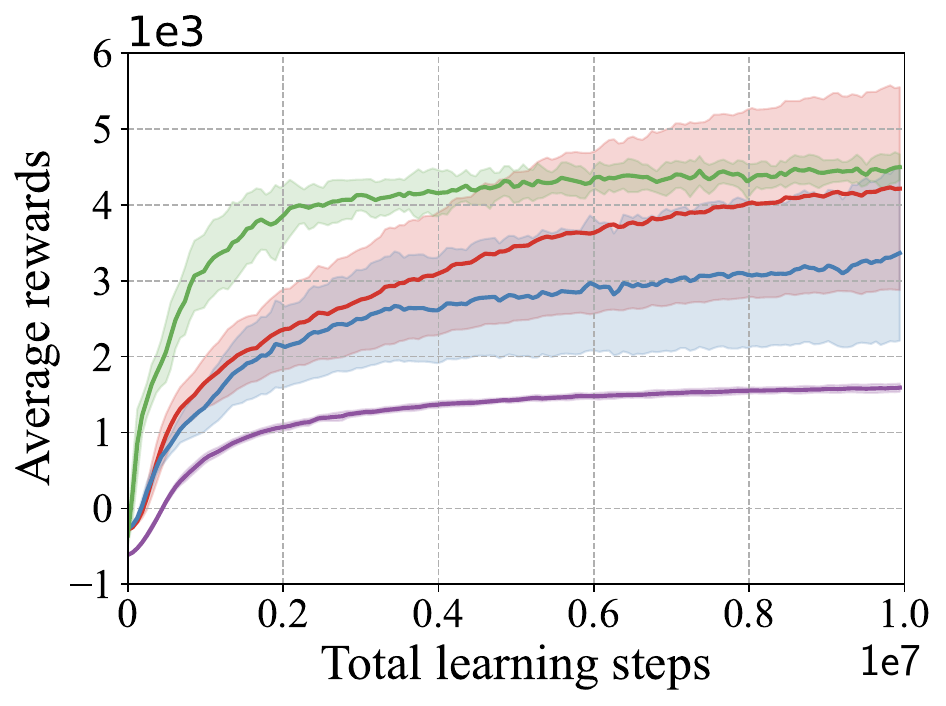}}
\subfigure[Humanoid-v1 + PPO]{\includegraphics[height=3.31cm]{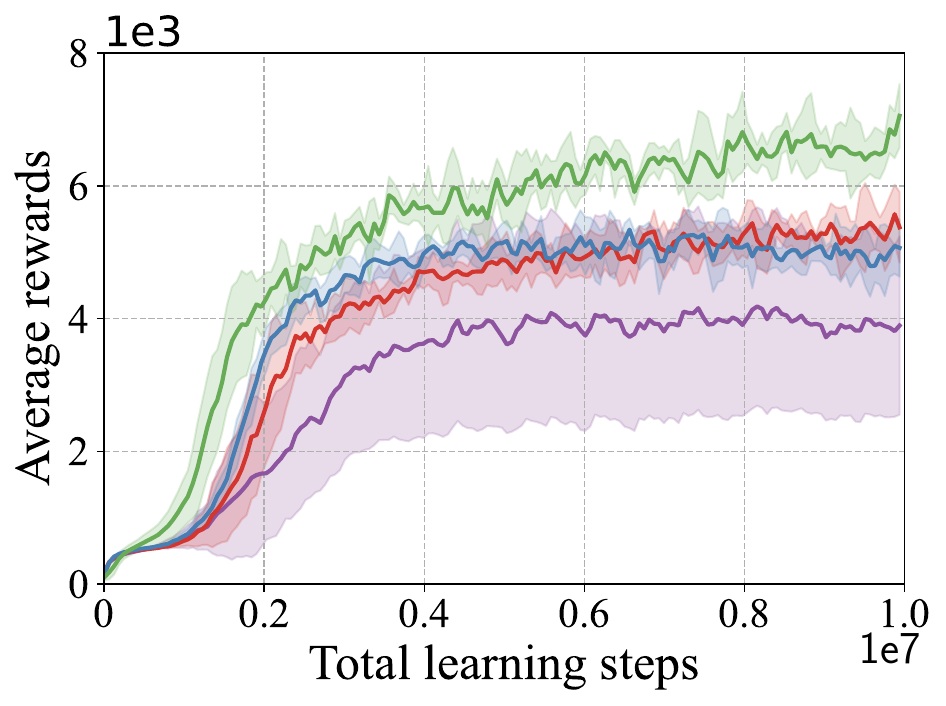}}
\subfigure[Humanoid-v3 + PPO]{\includegraphics[height=3.31cm]{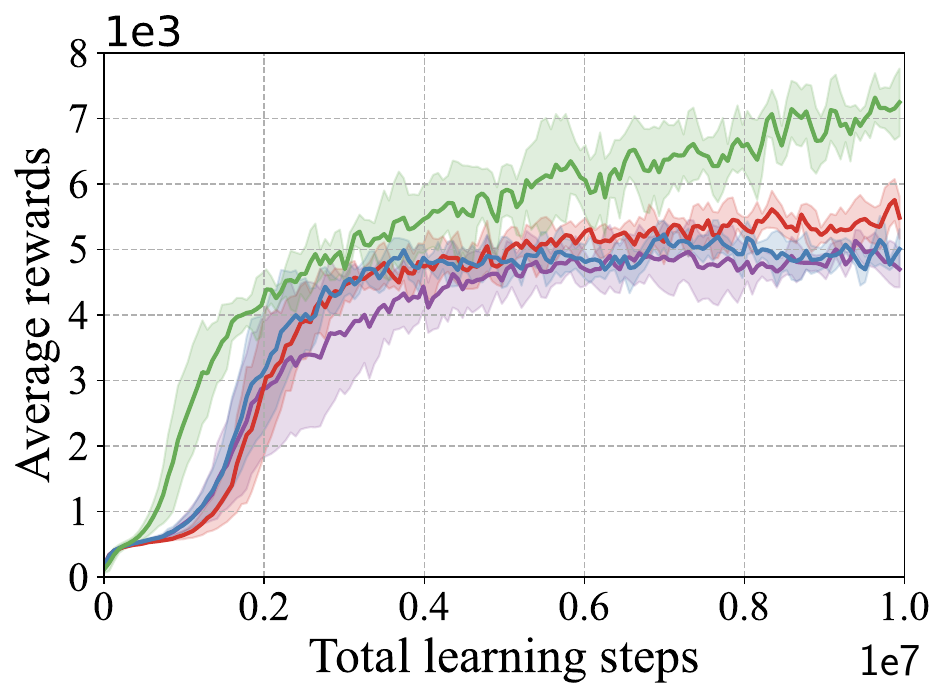}}
\subfigure[HumanoidStandup-v2 + PPO]{\includegraphics[height=3.31cm]{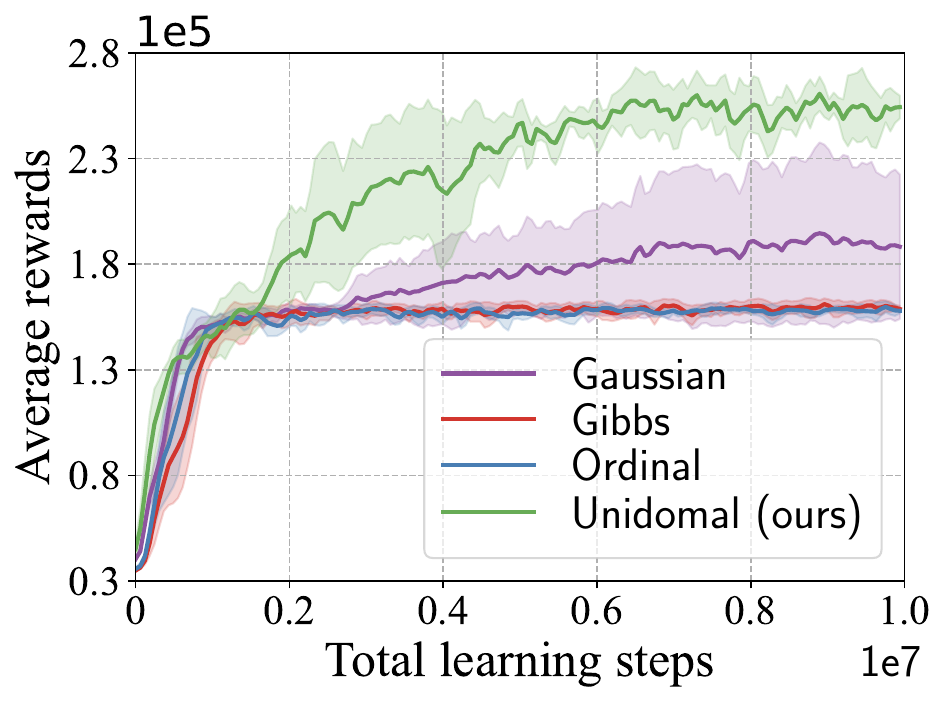}}
\caption{Learning curves of PPO with unimodal, discrete, ordinal and Gaussian policy on OpenAI gym MuJoCo locomotion tasks.
Solid lines and shadings denote the average values and standard deviation over $6$ random seeds.
All discrete policies have $K = 11$.
We see that the unimodal policy outperforms the other policies in terms of performance and stability on each task, especially on the Humanoid control tasks.}
\label{exp2}
\end{center}
\end{figure*}
\begin{figure*}[tb]
\begin{center}
\subfigure[Walker2D-v3]{\includegraphics[height=3.318cm]{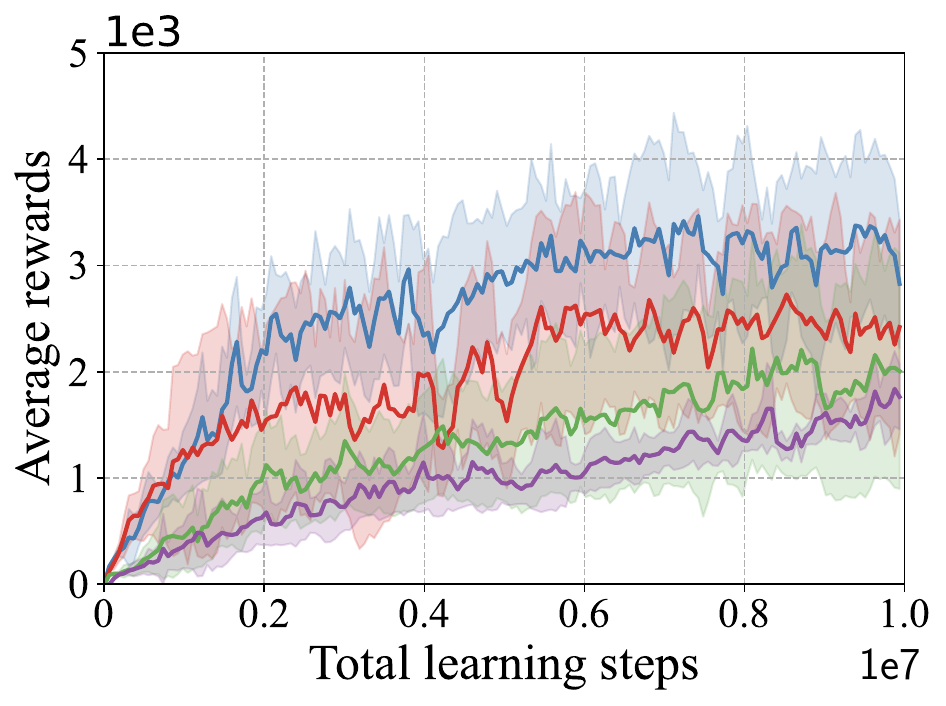}}
\subfigure[Swimmer-v3]{\includegraphics[height=3.318cm]{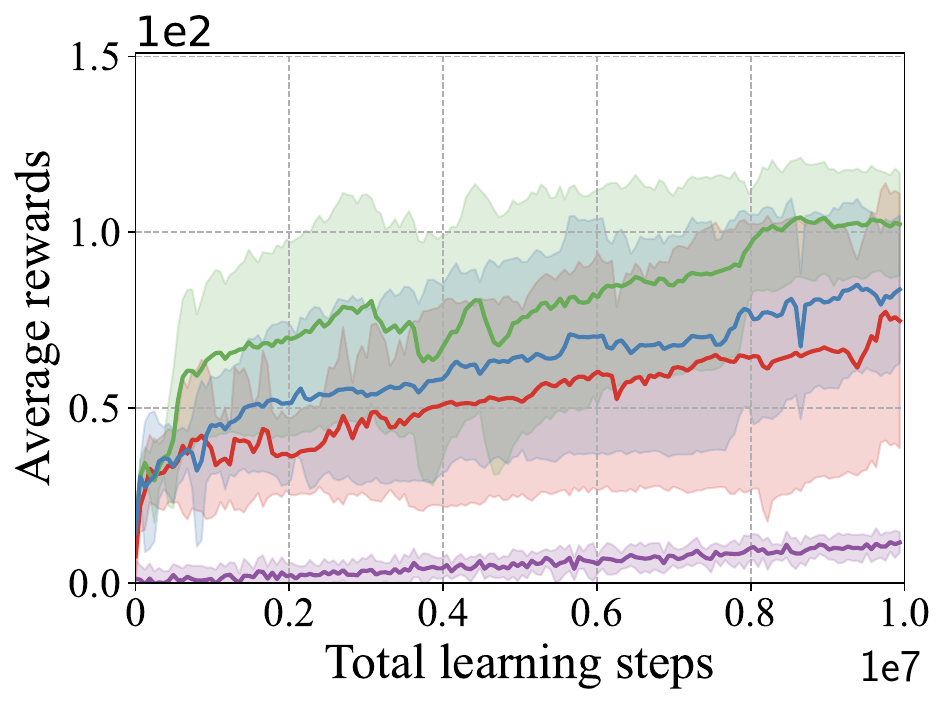}}
\subfigure[HalfCheetah-v3]{\includegraphics[height=3.318cm]{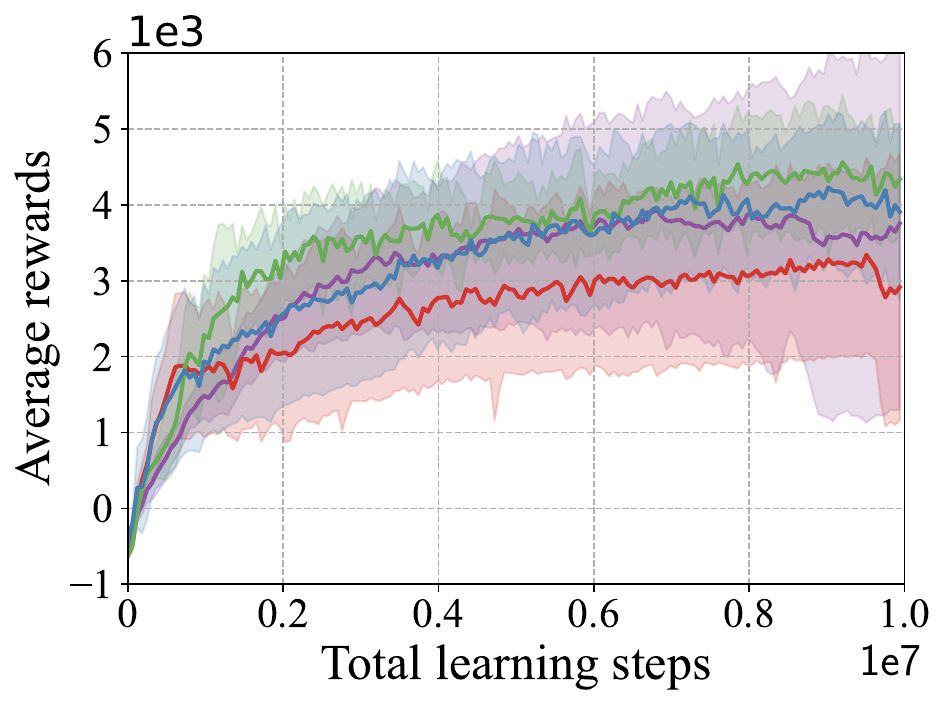}}
\subfigure[Humanoid-v3]{\includegraphics[height=3.318cm]{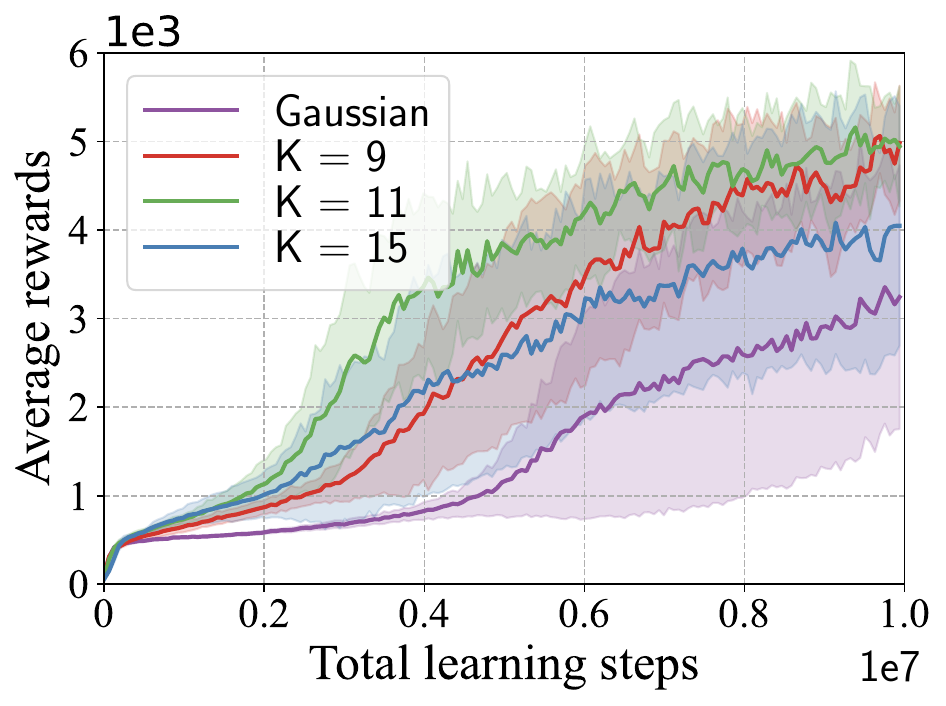}}
\caption{Performance as a function of the number of learning steps of ACKTR on OpenAI gym MuJoCo locomotion tasks.
Each curve corresponds to a different policy architecture (Gaussian or unimodal policy with varying bins $K = 9, 11, 15$).
Solid lines and shadings denote the average values and standard deviation over $6$ random seeds.
Our unimodal policy significantly outperforms the Gaussian policy on most tasks.}
\vskip -0.2in
\label{exp3}
\end{center}
\end{figure*}
\begin{figure}[tb]
\begin{center}
\subfigure[HalfCheetah-v3]{\includegraphics[height=3.24cm]{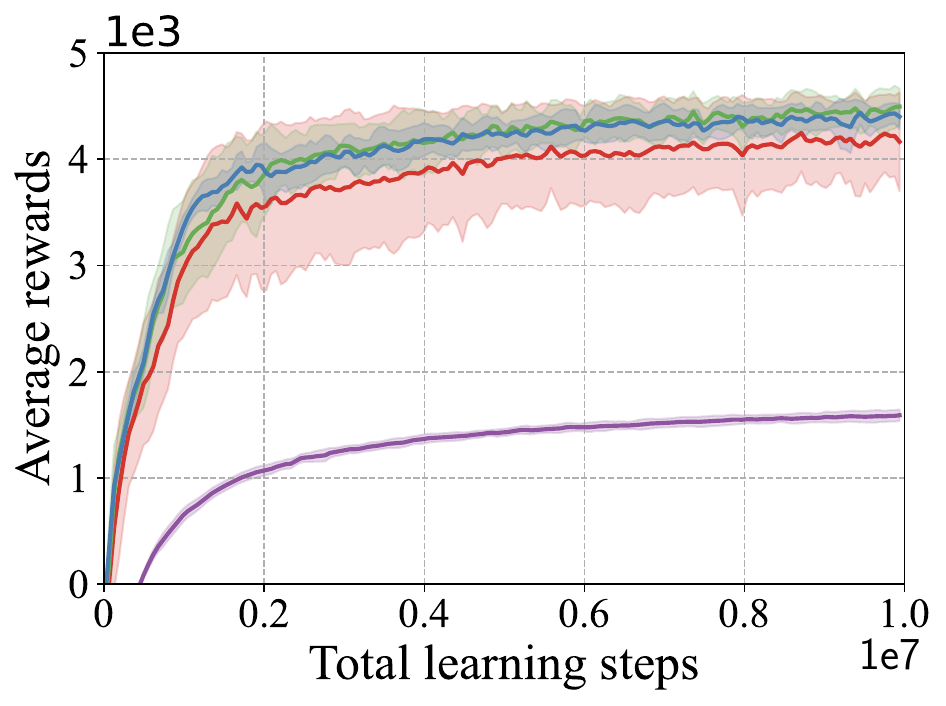}}
\subfigure[Humanoid-v2]{\includegraphics[height=3.24cm]{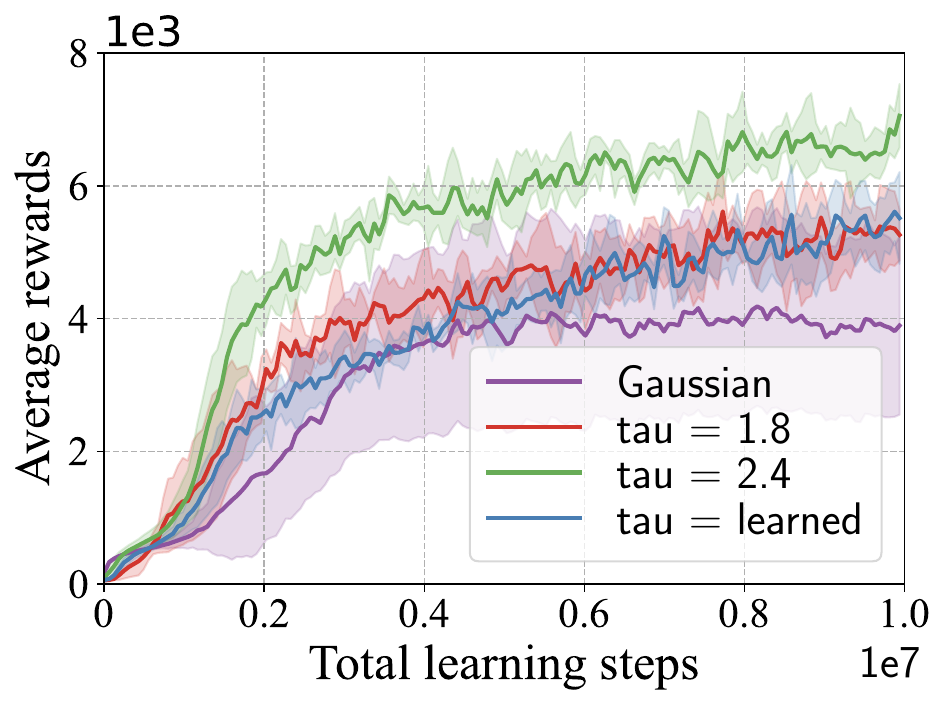}}
\caption{Performance comparison of PPO with the unimodal and Gaussian policy on two MuJoCo locomotion tasks in terms of temperature $\tau$ varies $1.8$, $2.4$, and learnable bias.
Each curve is averaged over $6$ random seeds and shows mean $\pm$ std performance.}
\vskip -0.22in
\label{exp4}
\end{center}
\end{figure}
\begin{figure}[tb]
\begin{center}
\subfigure[HalfCheetah-v2 + PPO]{\includegraphics[width=0.492\columnwidth]{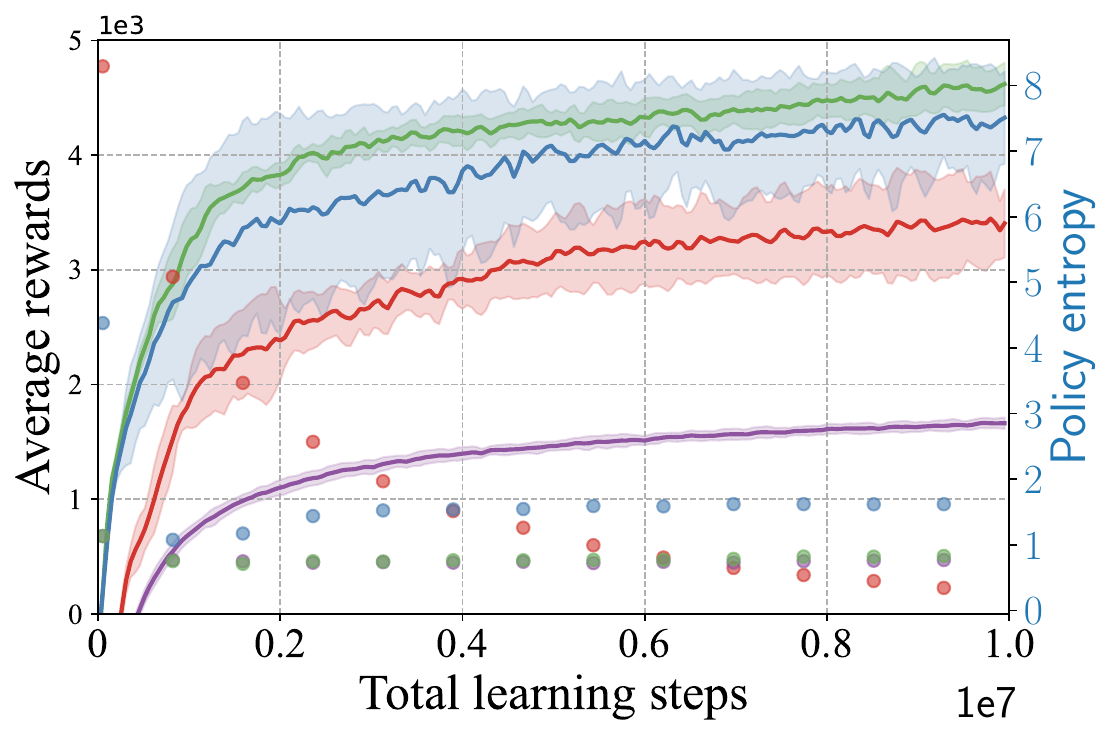}}
\subfigure[Ant-v3 + PPO]{\includegraphics[width=0.492\columnwidth]{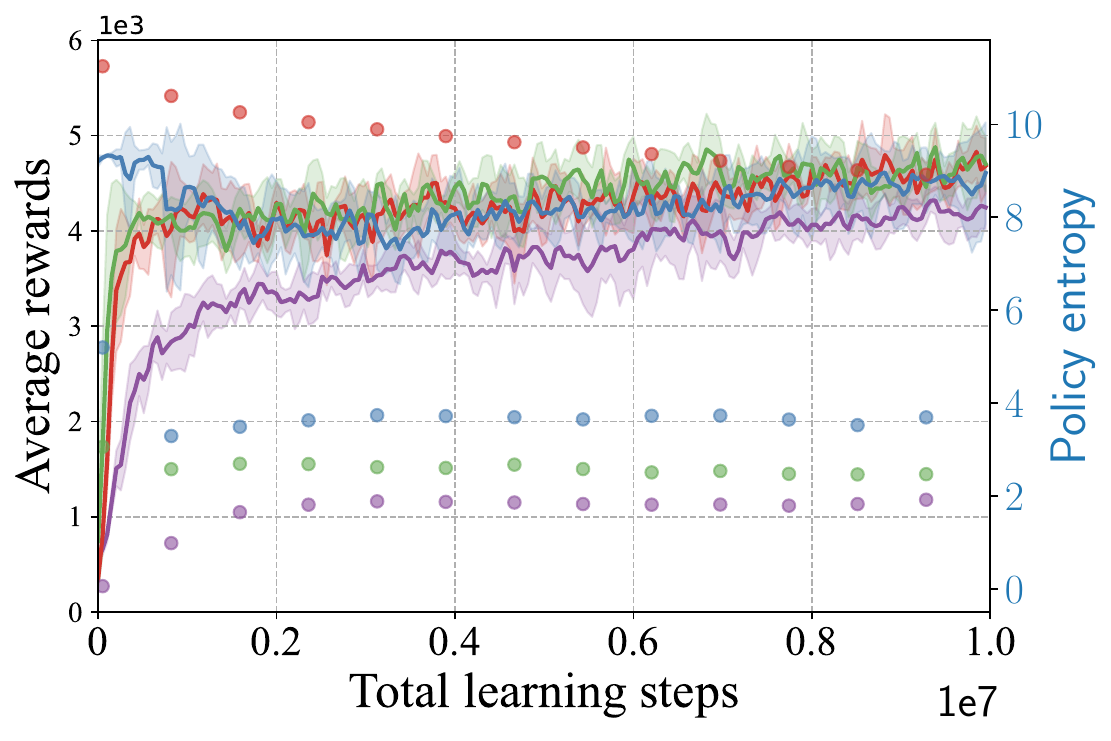}}
\subfigure[Humanoid-v3 + PPO]{\includegraphics[width=0.492\columnwidth]{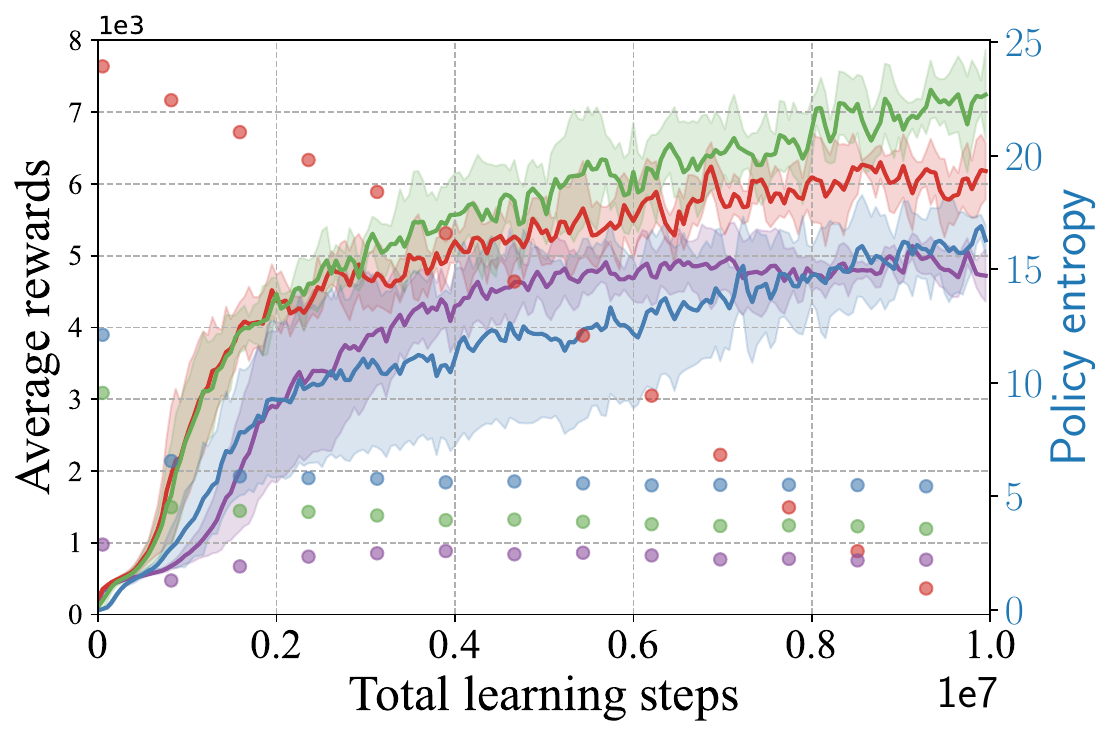}}
\subfigure[HumanoidStandup-v2 + PPO]{\includegraphics[width=0.492\columnwidth]{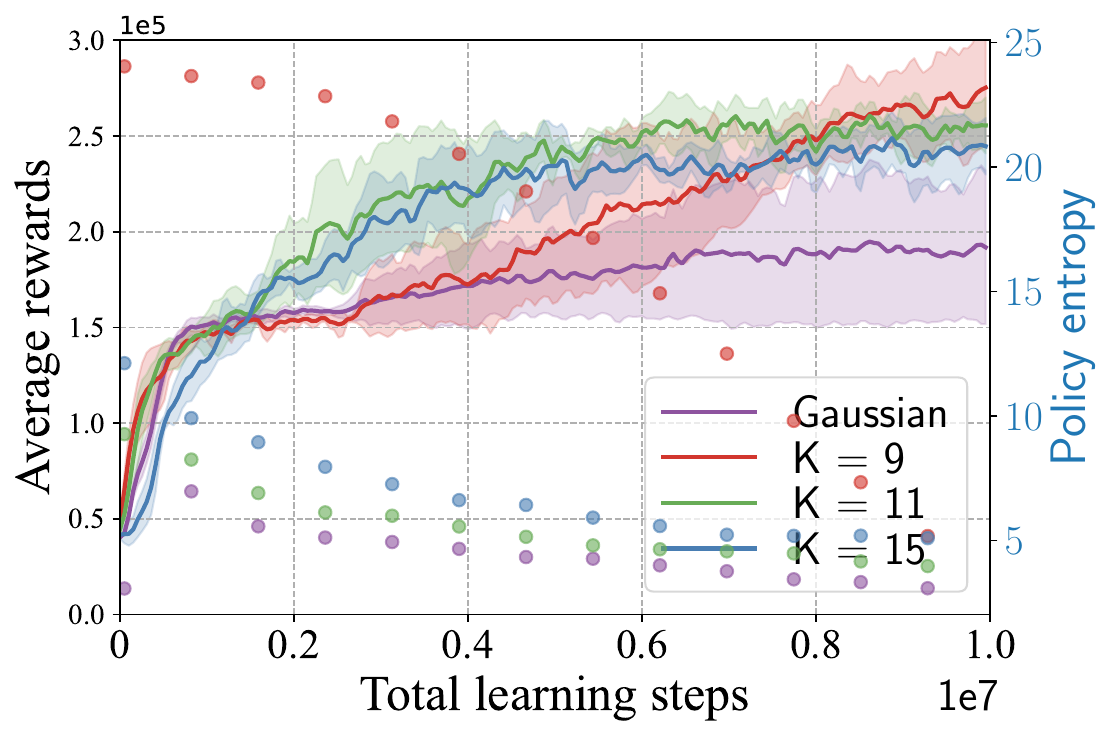}}

\caption{Learning curves and scatters of policy entropy for Gaussian and unimodal policies in continuous control tasks, which are represented shown in left- and right vertical axis, respectively.}
\vskip -0.22in
\label{exp5}
\end{center}
\end{figure}
\subsection{Comparison with Benchmark Baselines}
First, as a proof of concept, we compare the Gaussian policy with the unimodal policy (with varying $K$) in various MuJoCo control tasks using PPO and TRPO, as displayed in Fig.~\ref{exp1}.
The results show that, overall, PPO and TRPO with unimodal policy perform comparably to Gaussian policy on the easier tasks (e.g., Walker2D) and outperform them on the more complex tasks, both in terms of learning speed and the final performance (e.g., Humanoid).
PPO with unimodal policy also learns considerably faster and achieves more excellent performance than PPO with Gaussian policy. 
This implies that the Gaussian policy may suffer from unbounded distributions and require more samples to optimize its policy in more high-dimensional tasks.
For TRPO, the performance enhancements brought by the unimodal policy align consistently with PPO.
Both performance gains are most significant with PPO and TRPO, which should benefit from our more attentively designed distribution architecture.

Further, we compare straightforward architectural alternatives: Gaussian with tanh non-linearity as the output layer, and Beta distribution.
The primary rationale behind these architectures is their innate use of continuous unimodal probability distributions and their ability to confine sampled actions within the feasible range ($[-1, 1]$ for MuJoCo control tasks).
By construction, our proposed unimodal policy also bound the sampled actions within the feasible range using the right-truncated operation.
The mean and standard deviation results gathered across an extensive array of control tasks, are summarized in Table~\ref{texp1}.
In comparison to Gaussian and Gaussian with Tanh policies, the Beta policy yields higher scores, thereby mitigating the bias issue resulting from the infinite support mismatch of Gaussian distribution commonly employed in on-policy RL algorithms.
Our unimodal policy consistently outperforms the baselines across the majority of tasks. 
This outcome implies that the unimodal policy, with ordinal parameterizing distributions constrained by the Poisson distribution, can achieve notable performance by effectively considering the order of discrete actions.
Intriguingly, the unimodal policy, being a simplistic distribution, can attain such performance improvement in policy gradient methods.

Moreover, we also apply unimodal policy to another representative on-policy algorithm ACKTR with different atomic actions and compare it with the Gaussian policy on a set of MuJoCo tasks.
As shown in Fig.~\ref{exp3}, we find that ACKTR, with the unimodal policy, still outperforms its Gaussian approach by a large margin. 
It implies the superiority of our attentive unimodal architecture rather than larger networks.

Finally, we compare the performance of PPO with a unimodal policy against off-policy algorithms on highly complex Humanoid tasks (Humanoid-v2, Humanoid-v3, and HumanoidStandup-v2), including DDPG~\cite{lillicrap2015continuous}, SQL~\cite{haarnoja2017reinforcement}, SAC~\cite{haarnoja2018soft}, TD3~\cite{fujimoto2018addressing}, and DecQN~\cite{haarnoja2018soft}\footnote{For a fair comparison, we set the granularity of DecQN’s discretization to $11$ bins and the other hyperparameters settings are the same with DecQN~\cite{haarnoja2018soft}.}. 
Off-policy algorithms can potentially achieve better sample efficiency than on-policy algorithms by reusing samples, but for highly complex tasks such as Humanoid, even off-policy algorithms take many more samples to learn due to instability and less informative off-policy samples.
DecQN obtains higher competitive performance than the other baselines, which should benefit from the versatility of simple decoupled discrete control via leveraging a linear factorization over action dimensions.
In our experiments, PPO with unimodal policy achieves comparable or even better results than off-policy baselines, demonstrating its competitiveness in general complex applications. 
Our results show that PPO with unimodal policy is a promising method for complex control tasks, with performance comparable to state-of-the-art off-policy methods.

\subsection{Comparison with Discrete Policy}
In a similar vein to our work, the extension of discretizing continuous action space is applied to solve high-dimensional control tasks.
We evaluate the unimodal policy alongside the discrete and ordinal policy architectures on tasks with $K=11$.
As shown in Fig.~\ref{exp2}, the unimodal policy still significantly outperforms the Gibbs policy.
When compared to the ordinal policy, the unimodal policy learns a superior policy and achieves better performance due to the unimodal probability distribution.
Under a unimodal constraint, it is guaranteed that the two atomic actions on either side of the majority-selected action receive the next greatest amount of probability mass, which can result in a near-optimal policy during the early stages.
By this construction, the unimodal policy has fewer parameters than the discrete or ordinal policy when they share the same encoding architecture.
The results are predicted by our analysis showing that unimodal probability distributions of policy can still obtain high performance and better policies.
Although the Gibbs or ordinal policy can achieve enhanced performance, it is generally noisier without effectively utilizing the ordering information of discrete actions, relying more on the probability distribution of network output. 
By incorporating Poisson probability distribution constraints, the unimodal policy can more efficiently capture the continuity in the underlying continuous action space for generalization across discrete actions, especially in the high-dimension complex action space tasks.

\subsection{Stability and HyperParameter Analysis}
PG methods, being generally defined over parametric distribution functions, are susceptible to converging to sub-optimal extrema, where policy entropy is usually utilized to enhance exploration by preventing premature convergence to suboptimal policies. 
We assess the sensitivity of policy entropy and performance with varying numbers of atomic actions $K$ per dimension, presenting the results for PPO in Fig.~\ref{exp5}.
As shown in Fig.~\ref{exp5}, our proposed unimodal policy performs better with a larger number of atomic actions $K$ during the learning process.
With a larger $K$, we can execute more precise actions through fine discretization, ultimately reaching high-quality final policies. 
From the perspective of policy entropy, finer discretization enables more efficient exploration with larger entropy.
By examining both the performance and policy entropy in Fig.~\ref{exp5}, we observe that a higher absolute value of policy entropy yields better performance with more refined discretization, consistent with findings from~\cite{ahmed2019understanding}. 
Additionally, we also note that a moderate action bin $K$ (e.g., $9 \leq K \leq 15$) is sufficient for striking an appropriate balance between performance improvement and computational cost. 
The more minor variance of policy gradients tends to be more stable with finer discretization (larger $K$), potentially benefiting from the limited network output units inducing lower variance in the unimodal distributions.

Further, we study the performance of unimodal policy with varying temperature $\tau$ with $1.8$, $2.4$, and learnable bias, where the learnable $\tau$ is clipped to $[1.5, 3]$ as a bias.
As shown in Fig.~\ref{exp4}, our method can achieve robust performance under different temperatures $\tau$.
Note that a lower temperature $\tau$ could reduce exploration and encourage more deterministic behavior, which may lead to suboptimal policies without exposure to diverse environments.
The learned $\tau$ is more stable under different fixed temperatures $\tau$, though it does not show any significant gain over simply learning it as a bias.
One advantage of this technique is that the network can estimate its uncertainty on a per-example basis. 
Compared to their fixed temperature, it can better trade-off exploration versus exploitation under parameterized conditions.

\section{Conclusion and Limitations}\label{Sec5}
In this paper, we introduce a straightforward technique to enforce a unimodal ordinal probabilistic policy distribution using Poisson distributions. 
Given the inherent ordering between discretized actions in continuous control tasks, this property is crucial to consider in discrete stochastic policy gradients. 
Despite its simplicity, our approach significantly enhances the performance and stability of baseline algorithms, particularly in high-dimensional tasks with complex dynamics. 
Furthermore, the unimodal ordinal policy allows probability distributions to behave more sensibly in benchmark continuous control tasks.
For limitations, since we mainly aim to refine off-the-shelf on-policy RL algorithms, it is not sufficient to conclude that our method can transcend off-policy algorithms.
In a practical view, our unimodal policy is more suitable for improving policy for high-dimension control tasks in bounded time, while the current form does not provide an optimality guarantee.
For future work, we will focus on designing more attentive unimodal distributions and extend them to multi-agent reinforcement learning.

\bibliographystyle{IEEEtranN}
\renewcommand{\bibfont}{\small}
\bibliography{trans.bib}

\end{document}